\newif\ifpreprint

\preprinttrue

\ifpreprint
\documentclass[DTMColor]{ROB-New-Preprint}
\else
\documentclass[DTMColor]{ROB-New}
\fi

\usepackage{amssymb,amsfonts,amsmath,amscd}
\usepackage{graphicx} % for pdf, bitmapped graphics files
\usepackage{hyperref}
\usepackage{array}  % for table column width
\usepackage{xcolor}  % for colored text
\usepackage{subfigure}
\usepackage{acronym}
\usepackage{bm}
\usepackage{mleftright}
\usepackage{booktabs}
\usepackage{float}
\usepackage{adjustbox}
\newcolumntype{?}{!{\vrule width 0.2mm}}

\usepackage{tikz}
\usetikzlibrary{calc, arrows, shapes, positioning, decorations.pathreplacing,
	 calligraphy, shadings, angles, backgrounds, external}
\usepackage{pgfplots}
\pgfplotsset{compat=1.18}

\usepackage{pgfplots}
\usepackage{balance}

\acrodef{IMU}{inertial measurement unit}
\acrodefplural{IMU}{inertial measurement units}
\acrodef{GP}{Gaussian process}
\acrodefplural{GP}{Gaussian processes}
\acrodef{SDE}{stochastic differential equation}
\acrodefplural{SDE}{stochastic differential equations}
\acrodef{LTI}{linear time-invariant}
\acrodef{SLAM}{simultaneous localization and mapping}
\acrodef{NEES}{normalized estimation error squared}

\makeatletter
\renewcommand*\env@matrix[1][*\c@MaxMatrixCols c]{%
	\hskip -\arraycolsep
	\let\@ifnextchar\new@ifnextchar
	\array{#1}}
\makeatother

\gdef\p{\boldsymbol{\Phi}}
\gdef\q{\mathbf{Q}}
\gdef\a{\boldsymbol{\alpha}}
\gdef\dt{\Delta t}
\gdef\expontwo{e^{-2\alpha_i\dt_k}}
\gdef\expon{e^{-\alpha_i\dt_k}}
\gdef\w{\boldsymbol{\varpi}}
\gdef\x{\boldsymbol{\xi}}
\gdef\e{\boldsymbol{\epsilon}}
\gdef\J{\bm{\mathcal{J}}}

\newcommand{\change}[1]{{\color{black} #1}}

\begin{document}
	
\authormark{Keenan Burnett et al.}

\jnlPage{1}{21}
\jyear{2024}
\jdoi{10.1017/xxxxx}

\title{IMU as an Input vs. a Measurement of the State in Inertial-Aided State Estimation}

%\author{Keenan Burnett, Angela P. Schoellig, Timothy D. Barfoot}
\author[1]{Keenan Burnett\hyperlink{corr}{*}}
\address[1]{University of Toronto Institute for Aerospace Studies, Toronto, ON M3H5T6, Canada}

\author[2]{Angela P. Schoellig}
\address[2]{Technical University of Munich, 80333 Munich, Germany}

\author[1]{Timothy D. Barfoot}

\address{\hypertarget{corr}{*}Corresponding author. \email{keenan.burnett@robotics.utias.utoronto.ca}}

\ifpreprint
\articletype{}
\copytext{}
\received{}
\revised{}
\accepted{}
\else
\articletype{RESEARCH ARTICLE}
\received{xx xxx xxx}
\revised{xx xxx xxx}
\accepted{xx xxx xxx}
\fi

%%%%%%%%%%%%%%%%%%%%%%%%%%%%%%%%%%%%%%%%%%%%%%%%%%%%%%%%%%%%%%%%%%%%%%%%%%%%%%%%
\abstract{
	\change{

	 Treating IMU measurements as inputs to a motion model and then preintegrating these measurements has almost become a de-facto standard in many robotics applications. However, this approach has a few shortcomings. First, it conflates the IMU measurement noise with the underlying process noise. Second, it is unclear how the state will be propagated in the case of IMU measurement dropout. Third, it does not lend itself well to dealing with multiple high-rate sensors such as a lidar and an IMU or multiple asynchronous IMUs. In this paper, we compare treating an IMU as an input to a motion model against treating it as a measurement of the state in a continuous-time state estimation framework. We methodically compare the performance of these two approaches on a 1D simulation and show that they perform identically, assuming that each method's hyperparameters have been tuned on a training set. We also provide results for our continuous-time lidar-inertial odometry in simulation and on the Newer College Dataset. In simulation, our approach exceeds the performance of an imu-as-input baseline during highly aggressive motion. On the Newer College Dataset, we demonstrate state of the art results. These results show that continuous-time techniques and the treatment of the IMU as a measurement of the state are promising areas of further research. Code for our lidar-inertial odometry can be found at: \url{https://github.com/utiasASRL/steam_icp}.
	}
	}

\maketitle
\bibliographystyle{roblikecustom}

%%%%%%%%%%%%%%%%%%%%%%%%%%%%%%%%%%%%%%%%%%%%%%%%%%%%%%%%%%%%%%%%%%%%%%%%%%%%%%%%
\section{Introduction}

Inertial measurement units (IMUs) are important sensors in the context of state estimation. A popular approach in robotics is to treat IMU measurements as inputs to a motion model and then to numerically integrate the motion model to form relative motion factors between pairs of estimation times in a process known as preintegration \cite{lupton_tro12, forster_tro17, brossard_tro22}. In this paper, we investigate the treatment of IMUs as a measurement of the state using continuous-time state estimation with a Gaussian process motion prior. We compare treating an IMU as an input vs. a measurement on a simple 1D simulation problem. We then test our approach to lidar-inertial odometry using a simulated environment \change{and compare to a baseline that represents the IMU-as-input approach}. Finally, we provide experimental results for our lidar-inertial odometry on the Newer College Dataset.

Preintegration was initially devised as a method to avoid having to estimate the state at each measurement time in (sliding-window) batch trajectory estimation. We will show that by employing our continuous-time estimation techniques, we can achieve the same big-O complexity as classic preintegration, which is linear in the number of measurements. The contributions of this work are as follows:

\begin{itemize}
	\item We provide a detailed comparison of treating an IMU as an input to a motion model vs. as a measurement of the state on a 1D simulation problem. Such a comparison has not been previously presented in the literature.
	\item We show how to perform preintegration using heterogeneous factors (a combination of binary and unary factors) using continuous-time state estimation. To our knowledge, this has not been shown before in the literature.
	\item \change{We present our novel approach to lidar-inertial odometry using a Singer prior which includes body-centric acceleration in the state. We provide experimental results on the Newer College Dataset and on a custom simulated dataset. On the Newer College Dataset, we demonstrate state of the art performance.}
\end{itemize}

%Contributions: we show how to generalize preintegration to different types of measurements and additionally provide details on how to perform preintegration using continuous-time motion priors which has not been shown before in the literature.

%In robotics, we are often tasked with fusing multiple sensors in order to 

% It's taken as a given that sliding-window batch estimation is better than Kalman filtering.

\section{Related Work}

%A consequence of this is that the IMU input covariance $\mathbf{Q}_n$ has to tuned to each dataset in order to achieve a consistent estimator. We demonstrate this on a 1D simulation in section~\ref{section:sim_results}. 

Lupton and Sukkarieh \cite{lupton_tro12} were the first to show that a temporal window of inertial measurements could be summarized using a single relative motion factor in a method known as preintegration. Forster et al. \cite{forster_tro17} then showed how to perform preintegration on the manifold $SO(3)$. Subsequently, Brossard et al. \cite{brossard_tro22} demonstrated how to perform preintegration on the manifold of extended poses $SE_2(3)$, showing that this approach captures the uncertainty resulting from IMU measurements more consistently than $SO(3) \times \mathbb{R}^3$. All of these approaches treat the IMU as an input to a motion model. This approach has a few shortcomings. First, it conflates the IMU measurement noise with the underlying process noise. Second, it is unclear how the state and covariance should be propagate in the absence of IMU measurements. IMU measurement dropout is rare. However, it is worth considering the possibility in safety-critical applications. The third issue is that classic preintegration does not lend itself well to dealing with multiple high-rate sensors such as a lidar and an IMU or multiple \change{asyncrhonous} IMUs.

%Other continuous-time lidar-inertial odometry approaches include 

%

Previous work in continuous-time lidar-only odometry and lidar-inertial odometry include \cite{droeschel_icra18, quenzel_iros21, park_tro21, ng_iros21, lv_tmech23, lang_ral23} all of which employed B-splines. In B-spline approaches, exact derivatives of the continuous-time trajectory can be computed allowing for unary factors to be created for each accelerometer and gyroscope measurement, removing the need for preintegration. However, the spacing of control points has a large impact on the smoothness of B-spline trajectories. Determining this spacing is an important engineering challenge in B-spline approaches. This can be avoided by working with Gaussian processes instead. For a detailed comparison between B-splines and Gaussian processes in continuous-time state estimation, we refer the reader to Johnson et al. \cite{johnson_arxiv24}.

% For a detailed comparison between B-splines and Gaussian processes in continuous-time state estimation, we refer the reader to ...

% Other continuous-time lidar-inertial odometry approaches include ... which employ B-splines.

% 
Recently, Zheng and Zhu \cite{zheng_arxiv24} demonstrated continuous-time lidar-inertial odometry using Gaussian processes where rotation is decoupled from translation. They use a white-noise-on-jerk prior for position in a global frame, a white-noise-on-acceleration prior for rotation using a sequence of local Gaussian processes, and a white-noise-on-velocity prior for the IMU biases. Using this approach, they demonstrate competitive performance on the HILTI SLAM benchmark \cite{helmberger_ral22}. One consequence of estimating position in a global frame is that the power spectral density matrix $\mathbf{Q}$ of the prior must typically be isotropic whereas a body-centric approach allows for lateral-longitudinal-vertical dimensions to be weighted differently. In addition, as was shown by Brossard et al. \cite{brossard_tro22}, decoupling rotation from translation does not capture the uncertainty resulting from IMU measurements as accurately as keeping them coupled. However, one clear advantage of their approach is that all parts of the state are directly observable by the measurements, whereas in our approach angular acceleration is not directly observable.

In our prior work, we demonstrated continuous-time lidar-only odometry \cite{burnett_ral22} using a white-noise-on-acceleration motion prior. Lidar-only odometry using a white-noise-on-jerk prior \cite{tang_ral19} and a Singer prior \cite{wong_ral20b} have also been previously demonstrated. In this work, we choose to work with the Singer prior, which includes body-centric acceleration in the state. By including acceleration in the state, we can treat gyroscope and accelerometer measurements as measurements of the state rather than as inputs to a motion model.

\change{Another approach based on Gaussian processes is that of Le Gentil and Vidal-Calleja \cite{legentil_tro20, legentil_ijrr23}. They employ six independent Gaussian processes, three for acceleration in a global frame, and three for angular velocity. They optimize for the state at several inducing points given a set of IMU measurements over a preintegration window. They then analytically integrate these Gaussian processes to obtain preintegrated measurements that can be queried at any time of interest.}

\change{Prior lidar-inertial odometry methods include \cite{hu_robotica24, li_robotica24, xu_tro22, chen_icra23}. For a recent survey and comparison of open-source lidar-only and lidar-inertial odometry approaches, we refer the reader to \cite{fasiolo_robotica23}. For a more detailed literature review of lidar odometry, lidar-inertial odometry, and continuous-time state estimation, we refer the reader to our prior work \cite{burnett_arxiv24}. For a recent survey on continuous-time state estimation, we refer the reader to Talbot et al. \cite{talbot_arxiv24}. Extrinsic calibration of an IMU and calibration of the temporal offset between the IMU timestamps and the other sensors are both important areas of research. Recent works in this area include \cite{li_nie_robotica24, ferguson_robotica23}. By working with an existing dataset where extrinsic calibration and temporal synchronization has been taken care of, we can focus on the task of localization.
}

% Other continuous-time papers

% To our knowledge, our work is the first to provide a detailed 

\section{IMU as an Input vs. a Measurement}

\acused{IMU}

In this section, we investigate an approach where \ac{IMU} measurements are treated as direct measurements of the state using a continuous-time motion prior. We compare the performance of these two approaches on a simulated toy problem where we estimate the position and velocity of a 1D robot given noisy measurements of position and acceleration. Position measurements are acquired at a lower rate (10Hz) while the acceleration measurements are acquired at a higher rate (100Hz). For both approaches, our goal will be to reduce the number of states being estimated through preintegration.

%Preintegration is performed in a local frame such that the relative motion factor only needs to be computed once up front even as the linearization point changes.

%TODO: explain preintegration in one sentence.

\subsection{IMU as an Input} \label{sec:imu_input}

As a baseline, we consider treating IMU measurements as inputs to a discrete motion model, 
\begin{equation}
	 \underbrace{\begin{bmatrix} \mathbf{p}_{k} \\ \dot{\mathbf{p}}_{k} \end{bmatrix}}_{\mathbf{x}_{k}} =  \underbrace{\begin{bmatrix} \mathbf{1} & \Delta t_k \mathbf{1} \\ \mathbf{0} & \mathbf{1} \end{bmatrix}}_{\p(t_k, t_{k-1})} \underbrace{\begin{bmatrix} \mathbf{p}_{k-1} \\ \dot{\mathbf{p}}_{k-1} \end{bmatrix}}_{\mathbf{x}_{k-1}} + \underbrace{\begin{bmatrix} \frac{1}{2}\Delta t_k^2 \\ \Delta t_k \end{bmatrix}}_{\mathbf{B}_k} \mathbf{u}_k,
\end{equation} where $\Delta t_k = t_k - t_{k-1}$, $\p(t_k, t_{k-1})$ is the transition function, and $\mathbf{u}_k$ are acceleration measurements. A preintegration window is then defined between two endpoints, $(t_{k-1}, t_k)$, which includes times $\tau_0, \ldots, \tau_J$. Following the approach of \cite{lupton_tro12, forster_tro17}, the preintegrated measurements $\Delta \mathbf{x}_{k, k-1}$ are computed as
\begin{equation}
\Delta \mathbf{x}_{k,k-1} = \sum_{n=1}^{J} \p(\tau_J, \tau_n) \mathbf{B}_n \mathbf{u}_n.
\end{equation} These preintegrated measurements can be used to replace the acceleration measurements with a single relative motion factor between two endpoints:
\begin{subequations}
	\begin{align}
		J_{v,k} &= \frac{1}{2}\mathbf{e}_{k}^T \boldsymbol{\Sigma}_{k}^{-1} \mathbf{e}_{k},\\
		\mathbf{e}_{k} &= \mathbf{x}_k - \p(t_k, t_{k-1}) \mathbf{x}_{k-1} - \Delta \mathbf{x}_{k, k-1},		
	\end{align}
\end{subequations} where
\begin{equation}
\boldsymbol{\Sigma}_k = \sum_{n=1}^J \p(\tau_J, \tau_n) \mathbf{B}_n \mathbf{Q}_n \mathbf{B}_n^T \p(\tau_J, \tau_n)^T,
\end{equation} and $\mathbf{Q}_n$ is the covariance of the acceleration input $\mathbf{u}_n \sim \mathcal{N}(\mathbf{0}, \mathbf{Q}_n)$. Note that in this approach, uncertainty is propagated using the covariance on the acceleration input, $\mathbf{Q}_n$, which conflates IMU measurement noise and the underlying process noise. If the acceleration measurements were to drop out suddenly, it is unclear how the state and covariance should be propagated using this approach. It can be shown that preintegration is mathematically equivalent to marginalizing out the states between $\mathbf{x}_{k-1}$ and $\mathbf{x}_k$. The overall objective function that we seek to minimize is then
\begin{equation} \label{eq:objective0}
J(\mathbf{x}) = \sum_{k=0}^K \left(J_{v,k}(\mathbf{x}) + J_{y,k}(\mathbf{x})  \right),
\end{equation} where 
\begin{equation} \label{eq:unary_meas0}
J_{y,k} = \frac{1}{2} \left( \mathbf{y}_k - \mathbf{C}_k \mathbf{x}_k\right)^T \mathbf{R}_k^{-1} \left( \mathbf{y}_k - \mathbf{C}_k \mathbf{x}_k \right)
\end{equation} are the measurement factors, and $\mathbf{R}_k$ is the associated measurement covariance.

\subsection{Continuous-Time State Estimation}

In order to incorporate potentially asynchronous position and acceleration measurements, we propose to carry out continuous-time trajectory estimation as exactly sparse Gaussian process regression \cite{anderson_ar15, anderson_iros15, barfoot_ser17}. We consider systems with a \ac{GP} prior and a linear measurement model:
\begin{subequations}
\begin{align}
	\mathbf{x}(t) &\sim \mathcal{GP}(\check{\mathbf{x}}(t), \check{\mathbf{P}}(t, t^\prime)), \\
	\mathbf{y}_k &= \mathbf{C}_k \mathbf{x}(t_k) + \mathbf{n}_k,
\end{align}
\end{subequations} where $\mathbf{x}(t)$ is the state, $\check{\mathbf{x}}(t)$ is the mean function, $\check{\mathbf{P}}(t, t^\prime)$ is the covariance function, and $\mathbf{y}_k$ are measurements corrupted by zero-mean Gaussian noise $\mathbf{n}_k \sim \mathcal{N}(\mathbf{0}, \mathbf{R}_k)$. In this section, we restrict our attention to a class of GP priors resulting from \ac{LTI} \acp{SDE} of the form
\begin{align}
	\dot{\mathbf{x}}(t) &= \bm{A} \mathbf{x}(t) + \bm{B} \mathbf{u}(t) + \bm{L} \mathbf{w}(t), \\
	\mathbf{w}(t) &\sim \mathcal{GP}(\mathbf{0}, \bm{Q} \delta (t - t^\prime)), \nonumber
\end{align} where $\mathbf{w}(t)$ is a white-noise Gaussian process, $\bm{Q}$ is a power spectral density matrix, and $\mathbf{u}(t)$ is a known exogenous input. The general solution to this differential equation is
\begin{equation}
\mathbf{x}(t) = \p(t, t_0) \mathbf{x}(t_0) + \int_{t_0}^t \p(t, s) (\bm{B} \mathbf{u}(s) + \bm{L} \mathbf{w}(s)) ds,
\end{equation} where $\p(t, s) = \exp(\mathbf{A}(t - s))$ is the transition function. The mean function is
\begin{equation}
\check{\mathbf{x}}(t) = E[\mathbf{x}(t)] = \p(t, t_0) \check{\mathbf{x}}_0 + \int_{t_0}^t \p(t, s) \bm{B} \mathbf{u}(s) ds.
\end{equation} Over a sequence of estimation times, $t_0 < t_1 < \cdots < t_K$, the mean function can be written as
\begin{equation}
\check{\mathbf{x}}(t_k) = \p(t_k, t_0) \check{\mathbf{x}}_0 + \sum_{n=1}^k \p(t_k, t_n) \mathbf{B}_n \mathbf{u}_n,
\end{equation} assuming piecewise-constant input $\mathbf{u}_n$. This can be rewritten in a lifted form as
\begin{equation} \label{eq:prior_mean}
\check{\mathbf{x}} = \mathbf{A} \mathbf{B} \mathbf{u},
\end{equation} where $\mathbf{A}$ is the lifted lower-triangular transition matrix, the inverse of which is

\begin{equation} \label{eq:lifted_trans_inv}
\mathbf{A}^{-1} = \begin{bmatrix} 
	\mathbf{1} \\ 
	-\p(t_1,t_0)  & \ddots \\
	& \ddots &  \mathbf{1} \\
	& & -\p(t_K,t_{K-1}) & \mathbf{1}
\end{bmatrix},
\end{equation} $\mathbf{B} = \text{diag}(\mathbf{1}, \bm{B}_1, \cdots, \bm{B}_K)$, and $\mathbf{u} = [\check{\mathbf{x}}_0^T~\mathbf{u}_1^T~\cdots~\mathbf{u}_K^T]^T$. See \cite{barfoot_ser17} for further details on the formulations above. The covariance function is then
\begin{align}
	&\check{\mathbf{P}}(t, t^\prime) = E[(\mathbf{x}(t) - E[\mathbf{x}(t)]) (\mathbf{x}(t^\prime) - E[\mathbf{x}(t^\prime)])^T] \\
	&= \p(t, t_0) \check{\mathbf{P}}_0 \p(t^\prime, t_0)^T + \int_{t_0}^{\text{min}(t, t^\prime)} \p(t, s) \bm{L} \bm{Q} \bm{L}^T \p(t^\prime, s)^T ds \nonumber.
\end{align} The covariance can also be rewritten in a lifted form using the same set of estimation times as above,
\begin{equation} \label{eq:prior_cov}
\check{\mathbf{P}} = \mathbf{A}\mathbf{Q}\mathbf{A}^T,
\end{equation} where $\mathbf{Q} = \text{diag}(\check{\mathbf{P}}_0, \mathbf{Q}_1, \cdots, \mathbf{Q}_K)$, and
\begin{equation}
	\mathbf{Q}_k = \int_0^{\Delta t_k} \exp(\bm{A}(\Delta t_k - s)) \bm{L} \bm{Q} \bm{L}^T \exp(\bm{A}(\Delta t_k - s))^T ds.
\end{equation} Our prior over the entire trajectory can then be written as
\begin{equation} \label{eq:prior}
	\mathbf{x} \sim \mathcal{N}(\check{\mathbf{x}}, \check{\mathbf{P}}),
\end{equation} where $\check{\mathbf{P}}$ is the kernel matrix. Note that the inverse kernel matrix $\check{\mathbf{P}}^{-1}$ is block-tridiagonal thanks to the Markovian nature of the state. This sparsity property also holds for linear time-varying (LTV) \acp{SDE}, provided that they are also Markovian \cite{anderson_iros15}. The exact sparsity of $\check{\mathbf{P}}^{-1}$ is what allows us to perform efficient Gaussian process regression. This fact can be observed more easily by inspecting the following linear system of equations
\begin{equation}\label{eq:ct_gn}
\underbrace{\left( \check{\mathbf{P}}^{-1} + \mathbf{C}^T \mathbf{R}^{-1} \mathbf{C}  \right)}_{\hat{\mathbf{P}}^{-1}} \hat{\mathbf{x}} = \mathbf{A}^{-T} \mathbf{Q}^{-1} \mathbf{B} \mathbf{u} + \mathbf{C}^T \mathbf{R}^{-1} \mathbf{y},
\end{equation} where the Hessian is on the left-hand side, $\hat{\mathbf{P}}^{-1}$, is block-tridiagonal since $\check{\mathbf{P}}^{-1}$ is block-tridiagonal, and $\mathbf{C}^T \mathbf{R}^{-1} \mathbf{C}$ is block-diagonal. Thus, this linear system of equations can be solved in $O(K)$ time using a sparse Cholesky solver. The exact sparsity of $\check{\mathbf{P}}^{-1}$ also enables us to perform efficient Gaussian process interpolation. The standard \ac{GP} interpolation formulas are given by
\begin{subequations}\label{eq:interp}
	\begin{align}
		\hat{\mathbf{x}}(\tau) &= \check{\mathbf{x}}(\tau) + \check{\mathbf{P}}(\tau)\check{\mathbf{P}}^{-1} (\hat{\mathbf{x}} - \check{\mathbf{x}}),  \\
		\hat{\mathbf{P}}(\tau, \tau) &= \check{\mathbf{P}}(\tau, \tau) + \check{\mathbf{P}}(\tau)\check{\mathbf{P}}^{-1}\left(\hat{\mathbf{P}} - \check{\mathbf{P}} \right) \check{\mathbf{P}}^{-T} \check{\mathbf{P}}(\tau)^T, \vspace{-3mm}
	\end{align}
\end{subequations} where 
\begin{equation}
\check{\mathbf{P}}(\tau) = \begin{bmatrix} \check{\mathbf{P}}(\tau, t_0) & \check{\mathbf{P}}(\tau, t_1) & \cdots & \check{\mathbf{P}}(\tau, t_K) \end{bmatrix}.
\end{equation} The key to performing efficient interpolation relies on the sparsity of 
\begin{equation}
\check{\mathbf{P}}(\tau)\check{\mathbf{P}}^{-1} = \begin{bmatrix} \mathbf{0} & \cdots & \mathbf{0} & \boldsymbol{\Lambda}(\tau) & \boldsymbol{\Psi}(\tau) & \mathbf{0} & \cdots & \mathbf{0} \end{bmatrix}, \label{eq:p_sparse}
\end{equation} where 
\begin{subequations}
	\begin{align}
		\boldsymbol{\Psi}(\tau) &= \mathbf{Q}_\tau \boldsymbol{\Phi}(t_{k+1}, \tau)^T \mathbf{Q}_{k+1}^{-1}, \\
		\boldsymbol{\Lambda}(\tau) &= \boldsymbol{\Phi}(\tau, t_k) - \boldsymbol{\Psi}(\tau)  \boldsymbol{\Phi}(t_{k+1}, t_k),
	\end{align}
\end{subequations} are the only nonzero block-columns at indices $k+1$ and $k$, respectively. Thus, each interpolation query of the posterior trajectory is an $O(1)$ operation.

\begin{figure*} [t]
	\centering	
	\ifpreprint	
	\begin{tikzpicture}[
		line cap=round,
		edge/.style={>=latex, line width=1.4pt},
		transform/.style={>=latex, line width=1.0pt, dash pattern=on 2pt off 2pt},
		posenode/.style={draw, very thick, isosceles triangle,	isosceles triangle apex angle=45, minimum size=4mm, inner sep=1pt, outer sep=0pt},
		dot/.style={circle, fill, minimum size=#1, inner sep=0pt, outer sep=0pt}
		]
		\node[posenode, fill=white] (x0) {};
		\node[dot=5pt] (p0) [left=5mm of x0] {};
		\node[posenode, fill=white] (x1) [right=of x0] {};
		\node[posenode, fill=white] (x2) [right=of x1] {};
		\node[posenode, fill=white] (x3) [right=of x2] {};
		\node[posenode, fill=white] (x4) [right=of x3] {};
		\node[posenode, fill=white] (x5) [right=of x4] {};
		\node[posenode, fill=white] (x6) [right=of x5] {};
		\node[posenode, fill=white] (x7) [right=of x6] {};
		\node[posenode, fill=white] (x8) [right=of x7] {};
		
		\node[dot=5pt] (m0) [below=5mm of x0] {};
		\node[dot=5pt] (m1) [below=5mm of x1] {};
		\node[dot=5pt] (m2) [below=5mm of x2] {};
		\node[dot=5pt] (m3) [below=5mm of x3] {};
		\node[dot=5pt] (m4) [below=5mm of x4] {};
		\node[dot=5pt] (m5) [below=5mm of x5] {};
		\node[dot=5pt] (m6) [below=5mm of x6] {};
		\node[dot=5pt] (m7) [below=5mm of x7] {};
		\node[dot=5pt] (m8) [below=5mm of x8] {};
		
		\draw[-, edge] (m0.north) -- (x0.south);
		\draw[-, edge] (m1.north) -- (x1.south);
		\draw[-, edge] (m2.north) -- (x2.south);
		\draw[-, edge] (m3.north) -- (x3.south);
		\draw[-, edge] (m4.north) -- (x4.south);
		\draw[-, edge] (m5.north) -- (x5.south);
		\draw[-, edge] (m6.north) -- (x6.south);
		\draw[-, edge] (m7.north) -- (x7.south);
		\draw[-, edge] (m8.north) -- (x8.south);
		
		\node[] (t0) [above=0mm of x0] {$\mathbf{x}_0$};
		\node[] (t1) [above=0mm of x1] {$\mathbf{x}_1$};
		\node[] (t2) [above=0mm of x2] {$\mathbf{x}_2$};
		\node[] (t3) [above=0mm of x3] {$\mathbf{x}_3$};
		\node[] (t4) [above=0mm of x4] {$\mathbf{x}_4$};
		\node[] (t5) [above=0mm of x5] {$\mathbf{x}_5$};
		\node[] (t6) [above=0mm of x6] {$\mathbf{x}_6$};
		\node[] (t7) [above=0mm of x7] {$\mathbf{x}_7$};
		\node[] (t8) [above=0mm of x8] {$\mathbf{x}_8$};
		
		\draw[-, edge] (p0.east) -- (x0.west);
		\draw[-, edge] (x0.east) -- (x1.west) node [midway] (x10) {}; % node [midway, below] {$\phi^p$};
		\node[dot=5pt] at (x10) {};
		\draw[-, edge] (x1.east) -- (x2.west) node [midway] (x21) {}; % node [midway, below] {$\phi^p$};
		\node[dot=5pt] at (x21) {};
		\draw[-, edge] (x2.east) -- (x3.west) node [midway] (x32) {}; % node [midway, below] {$\phi^p$};
		\node[dot=5pt] at (x32) {};
		\draw[-, edge] (x3.east) -- (x4.west) node [midway] (x43) {}; % node [midway, below] {$\phi^p$};
		\node[dot=5pt] at (x43) {};
		\draw[-, edge] (x4.east) -- (x5.west) node [midway] (x54) {}; % node [midway, below] {$\phi^p$};
		\node[dot=5pt] at (x54) {};
		\draw[-, edge] (x5.east) -- (x6.west) node [midway] (x65) {}; % node [midway, below] {$\phi^p$};
		\node[dot=5pt] at (x65) {};
		\draw[-, edge] (x6.east) -- (x7.west) node [midway] (x76) {}; % node [midway, below] {$\phi^p$};
		\node[dot=5pt] at (x76) {};
		\draw[-, edge] (x7.east) -- (x8.west) node [midway] (x87) {}; % node [midway, below] {$\phi^p$};
		\node[dot=5pt] at (x87) {};
		
		\node[] (b1) [below=2mm of m1] {};
		\node[] (b3) [below=2mm of m3] {};
		\node[] (b5) [below=2mm of m5] {};
		\node[] (b7) [below=2mm of m7] {};
		
		\draw[-, decorate,
		decoration = {calligraphic brace}, ultra thick] (b3.north) -- (b1.north) node [midway, below] {marginalize};
		\draw[-, decorate,
		decoration = {calligraphic brace}, ultra thick] (b7.north) -- (b5.north) node [midway, below] {marginalize};
		
	\end{tikzpicture}
	\else
	\includegraphics{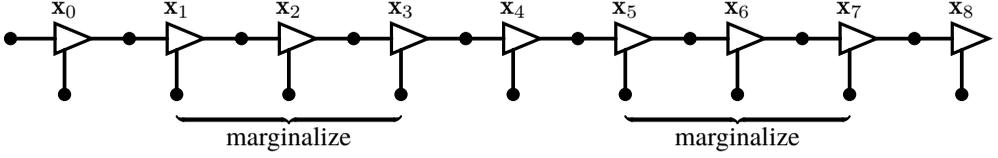}
	\fi
	\caption{In this factor graph, we consider a case where we would like to marginalize several states out of the full Bayesian posterior. The triangles represent states and the black dots represent factors. This factor graph could potentially correspond to doing continuous-time state estimation with binary motion prior factors, unary measurement factors, and a unary prior factor on the initial state $\mathbf{x}_0$}
	\label{fig:factor}
\end{figure*}

%... summarize quickly why interpolation is an O(1) operation.

%TODO: add more detail about the application of the GP stuff to the 1D problem using the notes below.
%
%Then, present and summarize results.
%
%Discuss training equations, chi-squared bounds, NEES test, why IMU-as-input is not consistent for this simulation, ergodic hypothesis, large amount of process noise vs. small IMU acceleration noise, cite Jeremy's paper.

\subsection{A Generalization to Preintegration} \label{sec:schur}

In section~\ref{sec:imu_input}, we showed how to perform preintegration when considering acceleration measurements as inputs to a motion model following closely from \cite{lupton_tro12, forster_tro17}. In this section, we generalize the concept of preintegration to support heterogeneous factors (a combination of binary factors and unary factors). As a motivating example, consider having measurements of position such as from a GPS and measurements of acceleration coming from an accelerometer. These are unary measurement factors, called such because they only involve a single state. The binary factors here are motion prior factors derived from our Gaussian process motion prior, called binary because they are between two states. In classic preintegration, the only factors that are preintegrated are binary factors. Here, we show that we can simply use the formulas for querying a Gaussian process posterior at the endpoints of the preintegration window to form a preintegration factor that summarizes all the measurements contained therein. First, we consider the joint density of the state at a set of query times $(\tau_0 < \tau_1 < \cdots < \tau_J)$ and the measurements,
\begin{equation}
	p\left(\begin{bmatrix} \mathbf{x}_\tau \\ \mathbf{y} \end{bmatrix} \right) = \mathcal{N} \left( \begin{bmatrix} \check{\mathbf{x}}_\tau \\ \mathbf{C} \check{\mathbf{x}}_\tau \end{bmatrix}, \begin{bmatrix} \check{\mathbf{P}}_{\tau, \tau} & \check{\mathbf{P}}_\tau \mathbf{C}^T \\ \mathbf{C} \check{\mathbf{P}}_\tau^T  & \mathbf{R} + \mathbf{C} \check{\mathbf{P}} \mathbf{C}^T \end{bmatrix}\right).
\end{equation} We then perform the usual factoring using a Schur complement to obtain the posterior,
\begin{align}
	&p(\mathbf{x}_\tau | \mathbf{y}) = \mathcal{N} \Big(  \check{\mathbf{x}}_\tau + \check{\mathbf{P}}_\tau \mathbf{C}^T (\mathbf{C} \check{\mathbf{P}} \mathbf{C} + \mathbf{R})^{-1} (\mathbf{y} - \mathbf{C} \check{\mathbf{x}}), \nonumber \\
	&~~~\check{\mathbf{P}}_{\tau, \tau} - \check{\mathbf{P}}_\tau \mathbf{C}^T(\mathbf{C} \check{\mathbf{P}} \mathbf{C}^T + \mathbf{R})^{-1} \mathbf{C} \check{\mathbf{P}}_\tau^T \Big),
\end{align} where we obtain expressions for $\hat{\mathbf{x}}_\tau$ and $\hat{\mathbf{P}}_{\tau, \tau}$. We rearrange this further by inserting $\check{\mathbf{P}}^{-1}\check{\mathbf{P}}$ after the first instance of $\check{\mathbf{P}}_\tau$ and by applying the Sherman-Morrison-Woodbury identities to obtain
\begin{subequations}
	\begin{align}
		&\hat{\mathbf{x}}_\tau = \check{\mathbf{x}}_\tau + \check{\mathbf{P}}_\tau \check{\mathbf{P}}^{-1} (\check{\mathbf{P}}^{-1} + \mathbf{C}^T\mathbf{R}^{-1}\mathbf{C})^{-1} \mathbf{C}^T \mathbf{R}^{-1} (\mathbf{y} - \mathbf{C} \check{\mathbf{x}}), \\
		&\hat{\mathbf{P}}_{\tau,\tau} = \check{\mathbf{P}}_{\tau,\tau} - \check{\mathbf{P}}_\tau \check{\mathbf{P}}^{-1} (\check{\mathbf{P}}^{-1} + \mathbf{C}^T\mathbf{R}^{-1}\mathbf{C})^{-1} \mathbf{C}^T \mathbf{R}^{-1} \mathbf{C} \check{\mathbf{P}}_\tau^T,
	\end{align}
\end{subequations} where we note that $(\check{\mathbf{P}}^{-1} + \mathbf{C}^T\mathbf{R}^{-1}\mathbf{C})$ is block-tridiagonal, and so the product $(\check{\mathbf{P}}^{-1} + \mathbf{C}^T\mathbf{R}^{-1}\mathbf{C})^{-1} \mathbf{C}^T \mathbf{R}^{-1} (\mathbf{y} - \mathbf{C} \check{\mathbf{x}})$ can be evaluated in $O(K)$ time using a sparse Cholesky solver. When we encounter products resembling $\mathbf{A}^{-1}\mathbf{b}$ where $\mathbf{A}$ is block-tridiagonal, we can instead solve $\mathbf{A}\mathbf{z} = \mathbf{b}$ for $\mathbf{z}$ using an efficient solver that takes advantage of the sparsity of $\mathbf{A}$. Finally, the product $\check{\mathbf{P}}_\tau \check{\mathbf{P}}^{-1}$ is quite sparse, having only two nonzero block columns per block row as shown earlier in \eqref{eq:p_sparse}. It follows that $\hat{\mathbf{x}}_\tau$ and $\hat{\mathbf{P}}_{\tau, \tau}$ can be computed in time that scales linearly with the number of measurements. Now, we consider the case where the query times consist of the beginning and end of a preintegration window, $(t_k, t_{k+1})$. The queried mean and covariance can be treated as pseudomeasurements summarizing the measurements contained in the preintegration window. We adjust our notation to make it clear that these are now being treated as measurements by using $\tilde{\mathbf{x}}$ instead of $\hat{\mathbf{x}}$. What we obtain is a joint Gaussian factor for the states at times $t_k$ and $t_{k+1}$,
\begin{align}  \label{eq:joint_gaussian_factor}
	J = \frac{1}{2} &\begin{bmatrix} \mathbf{x}_k - \tilde{\mathbf{x}}_k \\ \mathbf{x}_{k+1} - \tilde{\mathbf{x}}_{k+1} \end{bmatrix}^T \begin{bmatrix} \tilde{\mathbf{P}}_{k,k} & \tilde{\mathbf{P}}_{k,k+1} \\ \change{\tilde{\mathbf{P}}_{k,k+1}} & \tilde{\mathbf{P}}_{k+1,k+1} \end{bmatrix}^{-1} \begin{bmatrix} \mathbf{x}_k - \tilde{\mathbf{x}}_k \\ \mathbf{x}_{k+1} - \tilde{\mathbf{x}}_{k+1} \end{bmatrix}.
\end{align}

\begin{figure}[t]
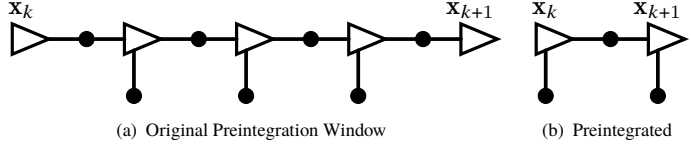

	\centering
	\subfigure[Original Preintegration Window]{

	\ifpreprint
		\begin{tikzpicture}[
			posenode/.style={draw, very thick, isosceles triangle,	isosceles triangle apex angle=45, minimum size=4mm, inner sep=1pt, outer sep=0pt},
			line cap=round,
			edge/.style={>=latex, line width=1.4pt},
			graphedge/.style={>=latex, line width=1pt},
			state/.style={draw, thick, line width=1pt, isosceles triangle,isosceles triangle apex angle=45, minimum size=4mm, inner sep=0pt, outer sep=0pt, fill=white},
			interpolated/.style={state, minimum size=2mm},
			dot/.style={draw, inner sep=0, circle, fill, minimum size=2mm, line width=0pt},
			mappoint/.style={dot, black, minimum size=1mm, outer sep=1mm},
			vfactor/.style={dot, green},
			pfactor/.style={dot, red},
			icpfactor/.style={dot, blue},
			every label/.style={align=left}
			]
			% main vertices and edges
	\node[posenode, fill=white] (x0) {};
	\node[posenode, fill=white] (x1) [right=of x0] {};
	\node[posenode, fill=white] (x2) [right=of x1] {};
	\node[posenode, fill=white] (x3) [right=of x2] {};
	\node[posenode, fill=white] (x4) [right=of x3] {};
	
	\node[dot=5pt] (m1) [below=5mm of x1] {};
	\node[dot=5pt] (m2) [below=5mm of x2] {};
	\node[dot=5pt] (m3) [below=5mm of x3] {};
	
	\draw[-, edge] (m1.north) -- (x1.south);
	\draw[-, edge] (m2.north) -- (x2.south);
	\draw[-, edge] (m3.north) -- (x3.south);
	
	\node[] (t0) [above=0mm of x0] {$\mathbf{x}_k$};
	\node[] (t4) [above=0mm of x4] {$\mathbf{x}_{k+1}$};
	
	\draw[-, edge] (x0.east) -- (x1.west) node [midway] (x10) {}; % node [midway, below] {$\phi^p$};
	\node[dot=5pt] at (x10) {};
	\draw[-, edge] (x1.east) -- (x2.west) node [midway] (x21) {}; % node [midway, below] {$\phi^p$};
	\node[dot=5pt] at (x21) {};
	\draw[-, edge] (x2.east) -- (x3.west) node [midway] (x32) {}; % node [midway, below] {$\phi^p$};
	\node[dot=5pt] at (x32) {};
	\draw[-, edge] (x3.east) -- (x4.west) node [midway] (x43) {}; % node [midway, below] {$\phi^p$};
	\node[dot=5pt] at (x43) {};
		\end{tikzpicture}
		
	\else
	\includegraphics{epsfigs/imu-figure2a.eps}
	\fi
	}
	\subfigure[Preintegrated]{
		\ifpreprint
			\begin{tikzpicture}[
			posenode/.style={draw, very thick, isosceles triangle,	isosceles triangle apex angle=45, minimum size=4mm, inner sep=1pt, outer sep=0pt},
			line cap=round,
			edge/.style={>=latex, line width=1.4pt},
			graphedge/.style={>=latex, line width=1pt},
			state/.style={draw, thick, line width=1pt, isosceles triangle,isosceles triangle apex angle=45, minimum size=4mm, inner sep=0pt, outer sep=0pt, fill=white},
			interpolated/.style={state, minimum size=2mm},
			dot/.style={draw, inner sep=0, circle, fill, minimum size=2mm, line width=0pt},
			mappoint/.style={dot, black, minimum size=1mm, outer sep=1mm},
			vfactor/.style={dot, green},
			pfactor/.style={dot, red},
			icpfactor/.style={dot, blue},
			every label/.style={align=left}
			]
			% main vertices and edges
			\node[posenode, fill=white] (x0) {};
			\node[posenode, fill=white] (x1) [right=of x0] {};
			
			\node[dot=5pt] (m0) [below=5mm of x0] {};
			\node[dot=5pt] (m1) [below=5mm of x1] {};
			
			\draw[-, edge] (m0.north) -- (x0.south);
			\draw[-, edge] (m1.north) -- (x1.south);
			
			\node[] (t0) [above=0mm of x0] {$\mathbf{x}_k$};
			\node[] (t1) [above=0mm of x1] {$\mathbf{x}_{k+1}$};
			\draw[-, edge] (x0.east) -- (x1.west) node [midway] (x10) {}; % node [midway, below] {$\phi^p$};
			\node[dot=5pt] at (x10) {};
		\end{tikzpicture}
		\else
		\includegraphics{epsfigs/imu-figure2b.eps}
		\fi
	}
	\caption{This figure depicts the results of our preintegration, which can incorporate heterogeneous factors. The resulting joint Gaussian factor in \eqref{eq:joint_gaussian_factor} can be thought of as two unary factors, one each for $\mathbf{x}_k$ and $\mathbf{x}_{k+1}$, and an additional binary factor between $\mathbf{x}_k$ and $\mathbf{x}_{k+1}$.}
	\label{fig:factor_comp}
\end{figure} \change{A diagram depicting how this affects the resulting factor graph is shown in Figure~\ref{fig:factor_comp}. Using this approach, we can \lq preintegrate' heterogeneous factors between pairs of states. One clear advantage of this approach is that it offers a tidy method for bookkeeping measurement costs and uncertainties. In the \ifpreprint appendix\else supplementary materials\fi, we provide an alternative formulation using a Schur complement with the same linear time complexity. Indeed, marginalization with a Schur complement is equivalent to the presented marginalization approach using a Gaussian process. However, the Gaussian process still serves a useful purpose in creating motion prior factors. Furthermore, the Gaussian process provides methods for interpolating the posterior.

It is unclear how to extend this marginalization approach to $SE(3)$ due to our choice to approximate $SE(3)$ trajectories using sequences of local Gaussian processes \cite{anderson_iros15}. It is possible that this marginalization approach could be applied using a global GP formulation such as the one presented by Le Gentil and Vidal-Calleja \cite{legentil_ijrr23}. However, their approach must contend with rotational singularities. We leave the extension to $SE(3)$ as an area of future work. In our implementation of lidar-inertial odometry, we instead use the posterior Gaussian process interpolation formula as in \eqref{eq:interp} to form continuous-time measurement factors. This is actually an approximation, as it is not exactly the same as marginalization. However, we have found the interpolation approach to work well in practice. Furthermore, the interpolation approach lends itself quite easily to parallelization enabling a highly efficient implementation.

}

\subsection{IMU as a Measurement}

%Note that in the GP approach, we are free to choose the estimation times. In order to make a direct comparison, we only estimate the state at the low-rate position measurements, the same as in preintegration.
%TODO: deal with u(t) being zero except for x0check. 
%TODO:  write covariance function
%TODO: talk about sparsity of inverse kernel matrix (also applies to LTV SDEs that are Markovian)
%TODO: talk about big-O complexity of solving trajectory estimation with GP
%TODO: Our GP prior is replaced by K - 1 binary motion prior factors
%TODO: we can simply treat the acceleration measurements as direct measurements of the state ... cost function ... compare to pre-integration.

%\begin{equation}
%\check{\mathbf{x}}(t_k) = \p(t_k, t_0) \check{\mathbf{x}}_0 + \sum_{n=1}^k \p(t_k, t_n)  
%\end{equation}

% In general, for a vector space variable, our prior takes the following form...

%Anderson and Barfoot \cite{anderson_iros15}. We begin by assuming a prior of the form

In our proposed approach, we treat IMU measurements as direct measurements of the state within a continuous-time estimation framework. For the 1D toy problem, we chose to use a Singer prior, which is defined by the following \acf{LTI} \acf{SDE},
\begin{align}
	\dddot{\mathbf{p}}(t) &= -\boldsymbol{\alpha} \ddot{\mathbf{p}}(t) + \mathbf{w}(t), \\
	\mathbf{w}(t) &\sim \mathcal{GP}(\mathbf{0}, \mathbf{Q}_c \delta (t - t^\prime)), \nonumber
\end{align} where $\mathbf{w}(t)$ is a white noise Gaussian process and $\mathbf{Q}_c~=~2 \boldsymbol{\alpha} \boldsymbol{\sigma}^2$ is the power spectral density matrix \cite{wong_ral20}. By varying $\boldsymbol{\alpha}$ \change{and $\boldsymbol{\sigma}^2$}, the Singer prior can model motion priors ranging from white-noise-on-acceleration \change{($\boldsymbol{\alpha} \rightarrow \boldsymbol{\infty}, \tilde{\boldsymbol{\sigma}}^2 = \boldsymbol{\alpha}^2 \boldsymbol{\sigma}^2$)} to white-noise-on-jerk ($\boldsymbol{\alpha} \change{\rightarrow} \mathbf{0}$). The discrete-time motion model is given by
\begin{align}
	&\mathbf{x}_{k} = \p(t_{k},t_{k-1}) \mathbf{x}_{k-1} + \mathbf{w}_k, \quad \mathbf{w}_k \sim \mathcal{N}(\mathbf{0}, \mathbf{Q}_k),
\end{align} where $\mathbf{w}_k$ is the process noise, $\mathbf{x}_k = [\mathbf{p}_{k}^T~\dot{\mathbf{p}}_{k}^T~\ddot{\mathbf{p}}_{k}^T]^T$,  $\p(t_{k},t_{k-1})$ is the state transition function, and $\mathbf{Q}_k$ is the discrete-time covariance. Expressions for $\p(t_{k},t_{k-1})$ and $\mathbf{Q}_k$ for the Singer prior are provided by Wong et al. \cite{wong_ral20} and are repeated in the \ifpreprint appendix\else supplementary materials\fi. The binary motion prior factors are given by
\begin{subequations}
	\begin{align}
		J_{v,k} &= \frac{1}{2}\mathbf{e}_{k}^T \mathbf{Q}_{k}^{-1} \mathbf{e}_{k}, \\
		\mathbf{e}_k &= \mathbf{x}_k - \p(t_k, t_{k-1}) \mathbf{x}_{k-1}.
	\end{align}
\end{subequations} The unary measurement factors have the same form as in \eqref{eq:unary_meas0} except that now accelerations are treated as direct measurements of the state. In addition, the overall objective is the same as in \eqref{eq:objective0}. We arrive at the linear system of equations from \eqref{eq:ct_gn}. By default, this approach would require that we estimate the state at each measurement time. In order to reduce the size of the state space, as in Section~\ref{sec:imu_input}, we build preintegrated factors using the approach presented in Section~\ref{sec:schur} in order to bundle together both the unary measurement factors as well as the binary motion prior factors into a single factor. In the exact same fashion as the IMU-as-input approach, we preintegrate between pairs of states associated with the low-rate measurement times.

\begin{figure}[t]
	\centering
	\ifpreprint
	\includegraphics[width=0.75\columnwidth]{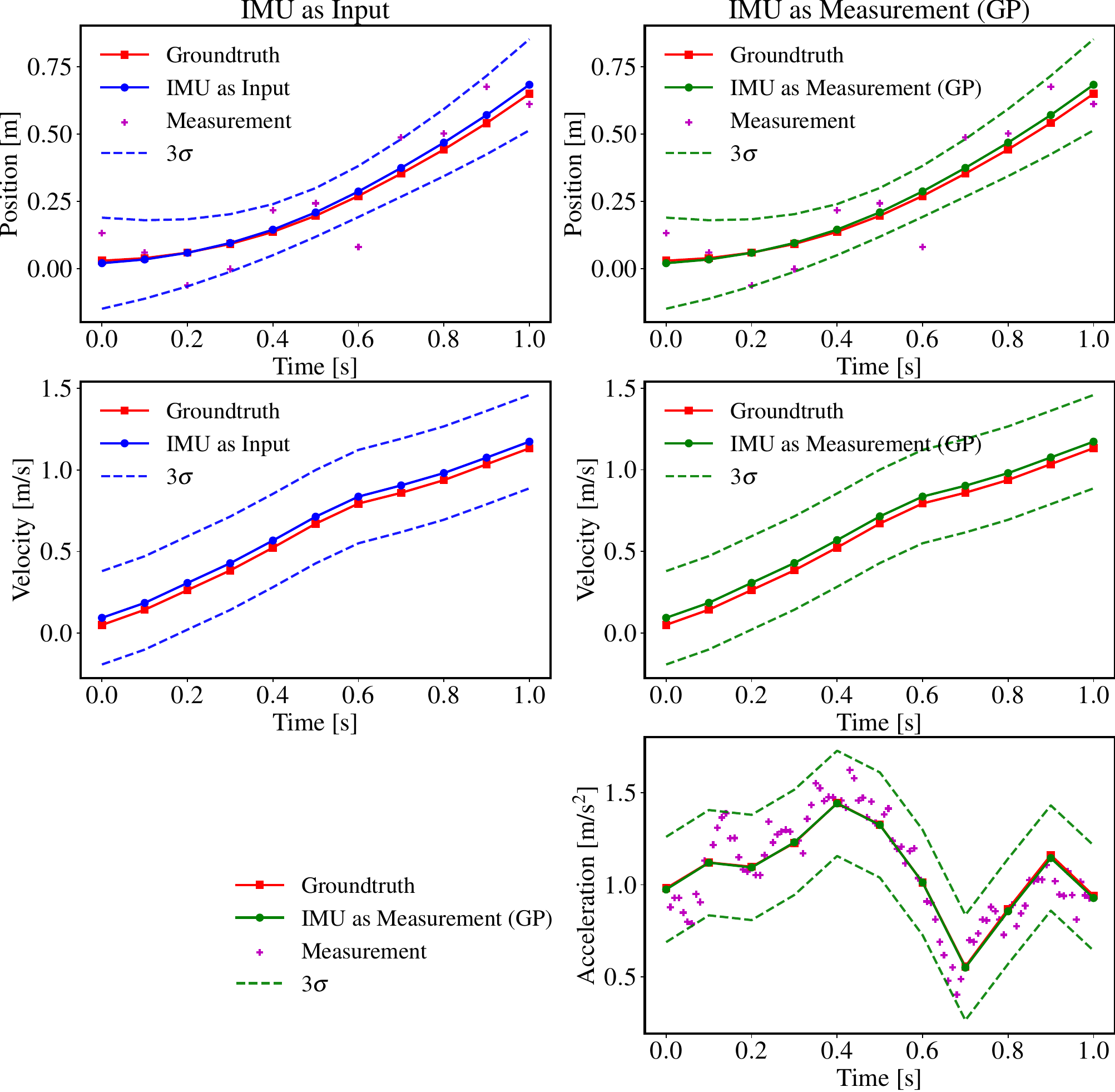}
	\else
	\includegraphics[width=0.75\columnwidth]{epsfigs/comp.eps}
	\fi
	\caption{The estimated trajectories of IMU-as-input and IMU-as-measurement are plotted alongside the ground-truth trajectory, which is sampled from white-noise-on-jerk motion prior with $Q_c = 1.0$. Both methods were pretrained on a hold-out validation set of simulated trajectories}
	\label{fig:comp}
\end{figure}

\begin{figure}[t]
	\centering
	\ifpreprint
	\includegraphics[width=0.35\columnwidth]{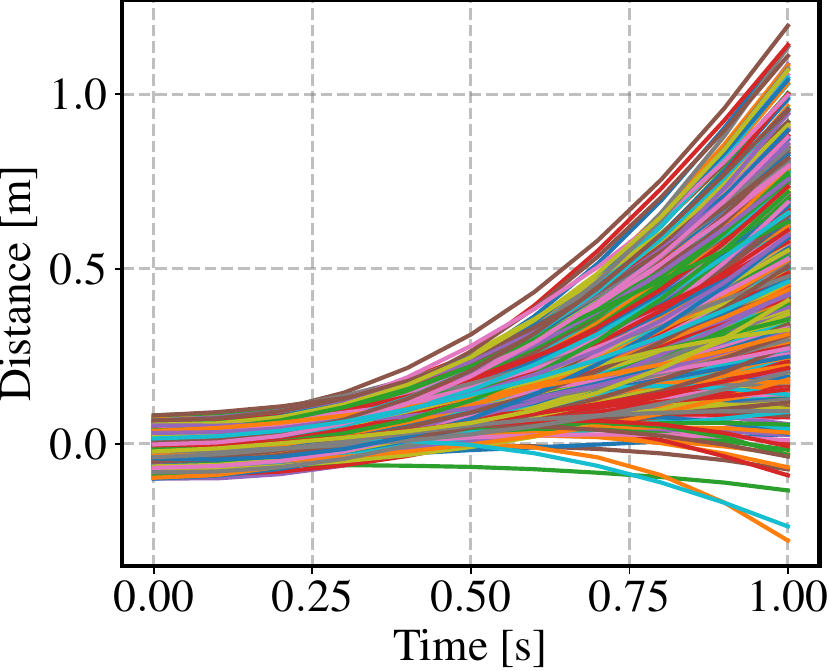}
	\else
	\includegraphics[width=0.35\columnwidth]{epsfigs/traj_ad_0.eps}
	\fi
	\caption{This figure depicts 1000 simulated trajectories sampled from a white-noise-on-jerk (WNOJ) prior where $\check{\mathbf{x}}_{0} = [0.0~0.0~1.0]^T$, $\check{\mathbf{P}}_{0} = \text{diag}(0.001, 0.001, 0.001)$, $Q_c = 1.0$}
	\label{fig:traj_ad_0}
\end{figure}

\subsection{Simulation Results} \label{section:sim_results}

In this section, we compare two different approaches to combining high-rate acceleration measurements with low-rate position measurements. We refer to these two different approaches as IMU-as-input and IMU-as-measurement. The comparison is conducted on a toy 1D simulation problem. We sample trajectories from a \ac{GP} motion prior as defined in \eqref{eq:prior} by using standard methods for sampling from a multidimensional Gaussian \cite{barfoot_ser17}.

In our first simulation, we sample trajectories from a white-noise-on-jerk (WNOJ) motion prior with $Q_c = 1.0$. Figure~\ref{fig:comp} provides a qualitative comparison of the IMU-as-input and IMU-as-measurement approaches for a single sampled simulation trajectory. Note that the performance of the two estimators including their 3-$\sigma$ confidence bounds appears to be nearly identical. Figure~\ref{fig:traj_ad_0} depicts 1000 trajectories sampled from this motion prior. In order to simulate noisy sensors, we also corrupt both position and acceleration measurements using zero-mean Gaussian noise where $y_{\text{pos},k} = \begin{bmatrix} 1 & 0 & 0 \end{bmatrix} \mathbf{x}_k + n_{\text{pos},k}$, $n_{\text{\text{pos}},k} \sim \mathcal{N}(0, \sigma^2_\text{pos})$, $y_{\text{acc},k} = \begin{bmatrix} 0 & 0 & 1 \end{bmatrix} \mathbf{x}_k + n_{\text{acc},k}$, $n_{\text{acc},k} \sim \mathcal{N}(0, \sigma^2_\text{acc})$. In the simulation, we set $\sigma_{\text{pos}} = 0.01m$, $\sigma_{\text{acc}} = 0.01m/s^2$.

\begin{figure}[t]
	\centering
	\ifpreprint
	\includegraphics[width=0.5\columnwidth]{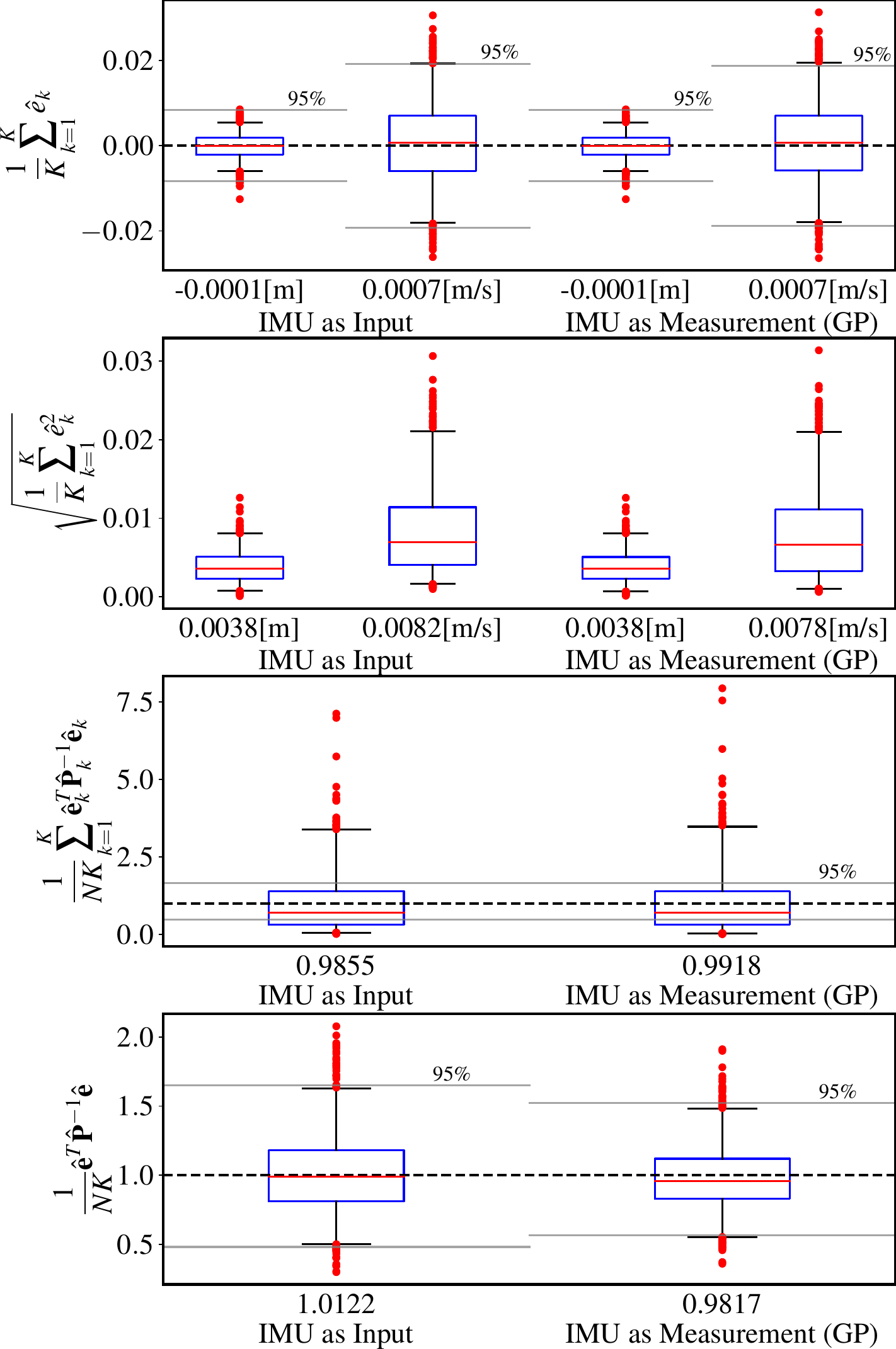}
	\else
	\includegraphics[width=0.5\columnwidth]{epsfigs/box_plot_ad_0.eps}
	\fi
	\caption{This figure compares the performance of the baseline IMU-as-input approach vs. our proposed IMU-as-measurement approach that leverages a \acf{GP} motion prior. The ground-truth trajectories are sampled from a white-noise-on-jerk (WNOJ) prior as shown in Figure~\ref{fig:traj_ad_0}. The IMU input covariance $Q_k$ for the IMU-as-input method was trained on a validation set, with a value of $0.00338$. The parameters of our proposed method was also trained on the same validation set, with resulting values of $\sigma^2 = 1.0069$, $\alpha = 0.0$. $R_{\text{pos}}$ for both methods was chosen to match the simulated noise added to the position measurements. Similarly, $R_{\text{acc}}$ for the IMU-as-measurement approach was set to match the simulated noise added to the acceleration measurements. In the bottom row, the chi-squared bounds have a different size because the dimension of the state in IMU-as-measurement is greater (it includes acceleration), and so the dimension of the chi-squared distribution increases, resulting in a tighter chi-squared bound}
	\label{fig:box_plot_ad_0}
\end{figure}

\begin{figure}[t]
	\centering
	\ifpreprint
	\includegraphics[width=0.35\columnwidth]{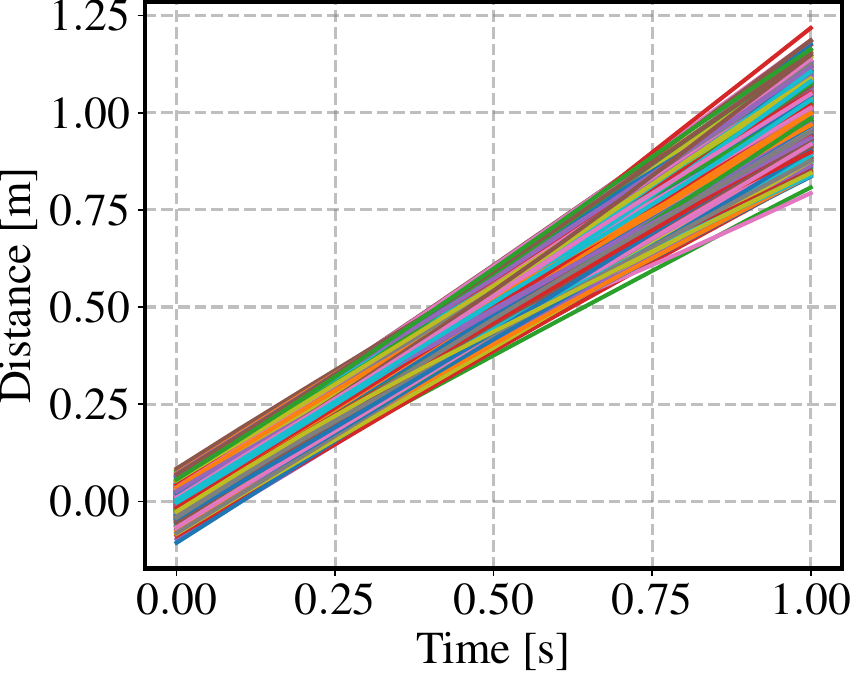}
	\else
	\includegraphics[width=0.35\columnwidth]{epsfigs/traj_ad_10.eps}
	\fi
	\caption{This figure depicts 1000 simulated trajectories sampled from a Singer prior where $\check{\mathbf{x}}_{0} = [0.0~1.0~0.0]^T$, $\check{\mathbf{P}}_{0} = \text{diag}(0.001, 0.001, 0.001)$, $\alpha = 10.0$, $\sigma^2 = 1.0$. A large value of $\alpha$ is intended to approximate a white-noise-on-acceleration (WNOA) prior}
	\label{fig:traj_ad_10}
\end{figure}

\begin{figure}[t]
	\centering
	\ifpreprint
	\includegraphics[width=0.5\columnwidth]{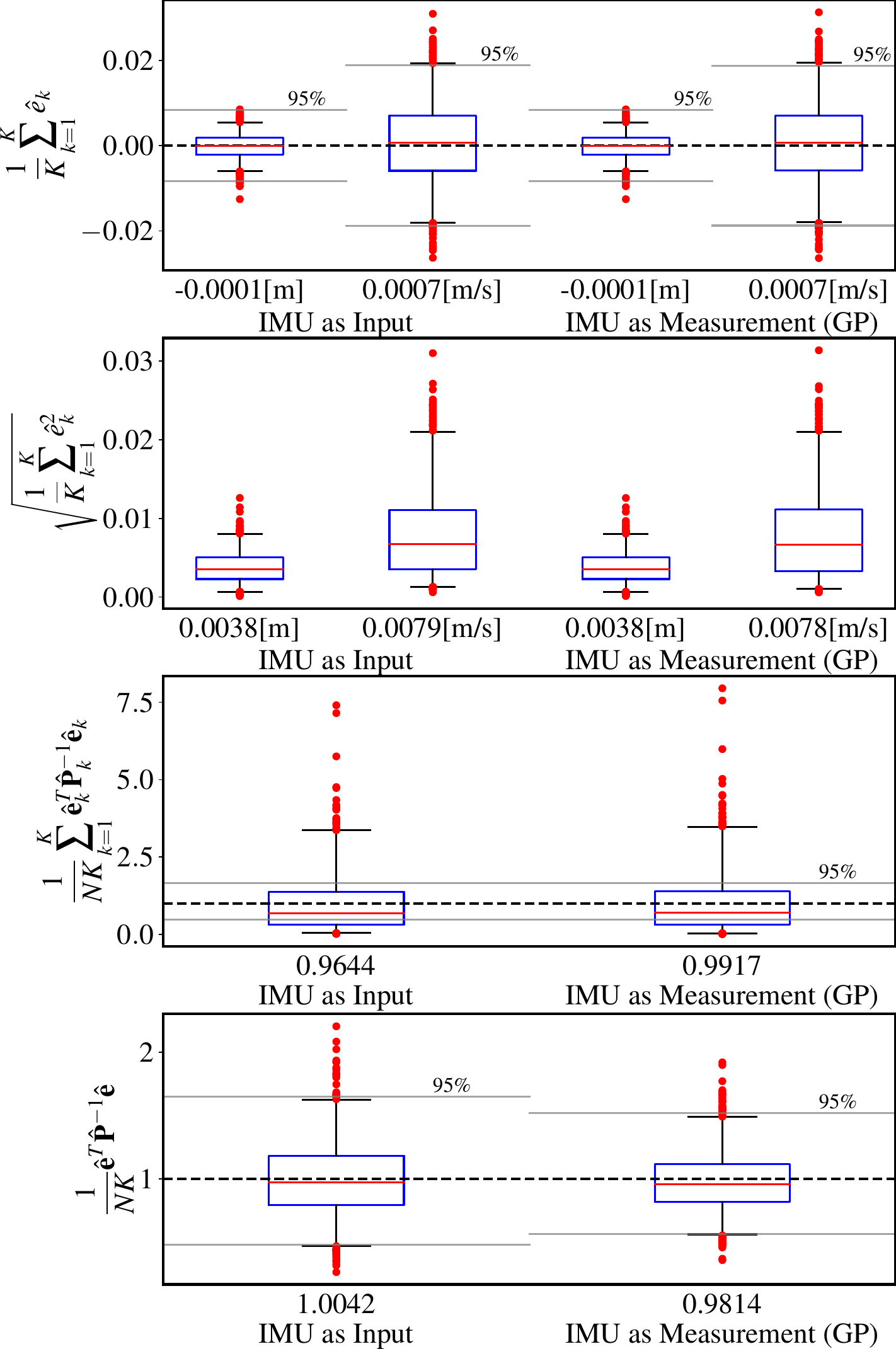}
	\else
	\includegraphics[width=0.5\columnwidth]{epsfigs/box_plot_ad_10.eps}
	\fi
	\caption{This figure summarizes the results of our second simulation experiment where the ground-truth trajectories were sampled from a Singer prior with $\alpha = 10.0$, $\sigma^2 = 1.0$. The trained acceleration input covariance for IMU-as-input was $Q_k \approx 0.00283$. The trained Singer prior parameters were $\alpha = 10.2442$, $\sigma^2 = 1.0074$. Note that even though the underlying prior changed by a lot from the first simulation to the second, from a white-noise-on-jerk prior to an approximation of a white-noise-on-acceleration prior, both estimators were able to remain unbiased and consistent}
	\label{fig:box_plot_ad_10}
\end{figure}

Both IMU-as-input and IMU-as-measurement have parameters that need to be tuned on the dataset. \change{For this purpose, we use a separate training set of 100 sampled trajectories.} For the IMU-as-input approach, we learn the covariance of the acceleration input $Q_k$ using maximum likelihood over the training set where we have the benefit of using noiseless ground truth states in simulation. The objective that we seek to minimize is \begin{equation}
J = \frac{1}{2}\ln\left| \check{\mathbf{P}} \right| +  \frac{1}{2T} \sum_{t=1}^T (\mathbf{x}_t - \check{\mathbf{x}})^T \check{\mathbf{P}}^{-1} (\mathbf{x}_t - \check{\mathbf{x}}), \label{eq:gp_factor}
\end{equation} where $\mathbf{x}_t$ are validation set trajectories, $\check{\mathbf{x}} = \mathbf{A} \mathbf{v}$ is the full trajectory prior for the IMU-as-input method,  $\check{\mathbf{P}} = \mathbf{A}\mathbf{Q}\mathbf{A}^T$ is the prior covariance over the whole trajectory, $\mathbf{A}$ is the lifted transition matrix as in \eqref{eq:lifted_trans_inv}, $\mathbf{v}~=~\begin{bmatrix} \check{\mathbf{x}}_0^T & \Delta \mathbf{x}_{1, 0}^T & \cdots & \Delta \mathbf{x}_{K, K-1}^T \end{bmatrix}^T$, and $\mathbf{Q}~=~\text{diag}(\check{\mathbf{P}}_0, \boldsymbol{\Sigma}_1, \cdots, \boldsymbol{\Sigma}_K)$.

Using a numerical optimizer, we found that $Q_k \approx 0.00338$ given a WNOJ motion prior with $Qc = 1.0$, and acceleration measurement noise of $\sigma^2_\text{acc} = 0.0001m^2/s^4$. Note that the input covariance $Q_k$ is clearly much greater than the simulated noise on the acceleration measurements $\sigma^2_\text{acc}$. This is because, for the IMU-as-input approach, the covariance on the acceleration input $Q_k$ is conflating two sources of noise, the IMU measurement noise and the underlying process noise. It also means that $Q_k$ has to be trained and adapted to new datasets in order to maintain a consistent estimator.

For the IMU-as-measurement approach, we need to train the parameters of our \acf{GP} motion prior. In order to highlight the versatility of our proposed approach, we employ a Singer motion prior \cite{wong_ral20}, which has the capacity to model priors ranging from white-noise-on-acceleration (WNOA) to white-noise-on-jerk (WNOJ). The Singer prior is parametrized by an inverse length scale matrix $\a$ and a variance $\boldsymbol{\sigma}^2$, both of which are diagonal. For values of $\a$ close to zero, numerical optimizers encounter difficulties due to numerical instabilities of $\mathbf{Q}_k$ and its Jacobians. Instead, we derive the analytical gradients to learn $\{\a, \boldsymbol{\sigma}^2\}$ using gradient descent following the approach presented by Wong et al. \cite{wong_ral20}. The objective that we seek to minimize is
\begin{equation} \label{eq:singer_obj}
J = \frac{1}{2} \sum_{t=1}^T \sum_{k=1}^K \left( \mathbf{e}_{k,t}^T \mathbf{Q}_{k,t}^{-1} \mathbf{e}_{k,t} + \ln \left| \mathbf{Q}_{k,t} \right| \right),
\end{equation} where both $\mathbf{e}_{k,t}$ and $\mathbf{Q}_{k,t}$ are functions of the Singer prior parameters $\{\a, \boldsymbol{\sigma}^2\}$. Further details on the analytical gradients are provided in the \ifpreprint appendix\else supplementary materials\fi. Note that the approach of Wong et al. \cite{wong_ral20} supports learning the parameters of the Singer prior even with noisy ground truth; however, it requires that we first estimate the measurement covariances and then keep them fixed during the optimization. In order to learn both the \ac{GP} parameters and the measurement covariances simultaneously, Wong et al. leverage exactly sparse Gaussian variational inference \cite{wong_ral20b}.

Figure~\ref{fig:box_plot_ad_0} shows the results of our first simulation experiment with the WNOJ prior. Each row in Figure~\ref{fig:box_plot_ad_0} is a box plot of a metric computed independently for each of the 1000 simulated trajectories. The blue boxes represent the interquartile range, the red lines are the medians, the whiskers correspond to the 2.5 and 97.5 percentiles, and the red dots denote outliers (data points beyond the whiskers). The black dashed lines in the first, third, and fourth rows corresponds to the target value that is 0 for the mean error and 1 for the \ac{NEES}. Underneath each box plot, we also compute the mean value of the metrics across all data points.

In the first row of Figure~\ref{fig:box_plot_ad_0}, we can see that the mean error in both position and velocity is close to zero for both estimators. The grey lines in the first row denote a 95\% two-sided confidence interval, a statistical test to check that the estimators are unbiased. We expect to see the whiskers of the box plots lie within the 95\% confidence interval in order to confirm that the estimator is unbiased, which is the case.

The third row displays a commonly used method for computing the \acf{NEES}. This method uses the marginals of the posterior covariance and relies on the ergodic hypothesis in order to treat the error from each timestep as being independent. In this case, we compute the marginal covariance at each timestep for position and velocity only so that the results of the two estimators can be compared directly. The grey lines denote a 95\% chi-squared bound, a statistical test for checking that the estimators are consistent. Interestingly, we observe that neither estimator passes the statistical test in this case even though the mean and median \ac{NEES} are close to 1. It appears that, in this case, the ergodic hypothesis is not valid.

In the fourth row, we present an alternative formulation of the \ac{NEES} that uses the full posterior covariance over the entire trajectory. In this case, we are satisfied to find that both estimators pass the statistical test for confirming that they are consistent. The main difference between this version of the \ac{NEES} and the previous one is that we have retained the cross-covariance terms between timesteps.

In summary, we observe that the two approaches achieve nearly identical performance. Both estimators are unbiased and consistent so long as their parameters are trained on a training set.

Figures~\ref{fig:traj_ad_10},~\ref{fig:box_plot_ad_10} depict the results of our second experiment where the ground-truth trajectories are sampled from a Singer prior with $\alpha = 10.0$, $\sigma^2 = 1.0$. This large value of $\alpha$ is intended to approximate a white-noise-on-acceleration prior. Our results show that both estimators are capable of adapting to a dataset with a different underlying motion prior while remaining unbiased and consistent.

\subsection{Discussion}

As mentioned previously, the big-O time complexity of classic preintegration and our approach are the same. In practice, our approach is slightly slower but not by much. Using a modern CPU, either approach can be considered real-time capable. Note that the number of preintegration windows could be adjusted and each preintegration window could be computed in parallel to make the approach more efficient. In this way, we could parallelize the solving of some estimation problems. We are motivated by sensor configurations that cannot be easily handled by classic preintegration such as multiple asynchronous high-rate sensors. This could include a lidar and an IMU or multiple \change{asynchronous} IMUs. 

\begin{figure*}[t]
	\centering
	\ifpreprint
	\begin{tikzpicture}[
		line cap=round,
		graphedge/.style={>=latex, line width=1pt},
		state/.style={draw, thick, line width=1pt, isosceles triangle,isosceles triangle apex angle=45, minimum size=4mm, inner sep=0pt, outer sep=0pt, fill=white},
		interpolated/.style={state, minimum size=2mm},
		dot/.style={draw, inner sep=0, circle, fill, minimum size=2mm, line width=0pt},
		mappoint/.style={dot, black, minimum size=1mm, outer sep=1mm},
		vfactor/.style={dot, green},
		preintfactor/.style={dot, magenta},
		pfactor/.style={dot, red},
		icpfactor/.style={dot, blue},
		every label/.style={align=left}
		]
		
		% main vertices and edges
		\node[] (p1) {};
		\node[] (p2) [right= of p1.center] {};
		% initial link
		\draw[graphedge, dotted] (p1) -- (p2);
		\begin{scope}
			\clip (p2) circle (6mm);
			\draw[graphedge] (p1) -- (p2);
		\end{scope}
		
		\def \dx{0mm};
		\def \vy{0.7};
		% state triangles
		\node[state, fill=gray, label=above left:{$\mathbf{x}_{k-2}$}] (s0) [right = of p1.center] {};
		\node[interpolated] (s1) [right =of s0, xshift=\dx] {};
		\node[interpolated] (s2) [right =of s1, xshift=\dx] {};
		\node[interpolated] (s3) [right =of s2, xshift=\dx] {};
		\node[state] (s4) [right =of s3, xshift=\dx] {};
		\node[interpolated] (s5) [right =of s4, xshift=\dx] {};
		\node[interpolated] (s6) [right =of s5, xshift=\dx] {};
		\node[interpolated] (s7) [right =of s6, xshift=\dx] {};
		\node[state, label=above right:{$\mathbf{x}_{k}$}] (s8) [right =of s7, xshift=\dx] {};
		
		% Prior factors
		\draw[graphedge] (s0.east) edge node[midway, pfactor] {} (s1.west);
		\draw[graphedge] (s1.east) edge node[midway, pfactor] {} (s2.west);
		\draw[graphedge] (s2.east) edge node[midway, pfactor] {} (s3.west);
		\draw[graphedge] (s3.east) edge node[midway, pfactor] {} (s4.west);
		\draw[graphedge] (s4.east) edge node[midway, pfactor] {} (s5.west);
		\draw[graphedge] (s5.east) edge node[midway, pfactor] {} (s6.west);
		\draw[graphedge] (s6.east) edge node[midway, pfactor] {} (s7.west);
		\draw[graphedge] (s7.east) edge node[midway, pfactor] {} (s8.west);
		
		% sliding window prior
		\node[pfactor] (v0) at ($(s0.center) + (0, \vy)$) {};
		
		% velocity factor
		\node[vfactor] (v1) at ($(s1.center) + (0, \vy)$) {};
		\node[vfactor] (v2) at ($(s2.center) + (0, \vy)$) {};
		\node[vfactor] (v3) at ($(s3.center) + (0, \vy)$) {};
		\node[vfactor] (v4) at ($(s4.center) + (0, \vy)$) {};
		\node[label={$\mathbf{x}_{k-1}$}]  (ttt) at ($(s4.center) + (0, \vy)$) {};
		\node[vfactor] (v5) at ($(s5.center) + (0, \vy)$) {};
		\node[vfactor] (v6) at ($(s6.center) + (0, \vy)$) {};
		\node[vfactor] (v7) at ($(s7.center) + (0, \vy)$) {};
		\node[vfactor] (v8) at ($(s8.center) + (0, \vy)$) {};
		
		\draw[graphedge] (s0) -- (v0);
		\draw[graphedge] (s1) -- (v1);
		\draw[graphedge] (s2) -- (v2);
		\draw[graphedge] (s3) -- (v3);
		\draw[graphedge] (s4) -- (v4);
		\draw[graphedge] (s5) -- (v5);
		\draw[graphedge] (s6) -- (v6);
		\draw[graphedge] (s7) -- (v7);
		\draw[graphedge] (s8) -- (v8);
		
		% local map points
		\node[mappoint] (mm1) at ($(s0.center) + (1mm, -11mm)$) {};
		\node[mappoint] (mm2) at ($(s0.center) + (4mm, -13mm)$) {};
		\node[mappoint] (mm3) at ($(s0.center) + (4mm, -15mm)$) {};
		\node[mappoint] (mm4) at ($(s0.center) + (6mm, -17mm)$) {};
		\node[mappoint] (mm5) at ($(s0.center) + (8mm, -19mm)$) {};
		\node[mappoint] (mm6) at ($(s0.center) + (8mm, -21mm)$) {};
		\node[mappoint] (mm7) at ($(s0.center) + (6mm, -24mm)$) {};
		\node[mappoint] (mm8) at ($(s0.center) + (8mm, -27mm)$) {};
		\node[mappoint] (mm11) at ($(s0.center) + (-2mm, -11mm)$) {};
		\node[mappoint] (mm12) at ($(s0.center) + (0mm, -13mm)$) {};
		\node[mappoint] (mm13) at ($(s0.center) + (2mm, -16mm)$) {};
		\node[mappoint] (mm14) at ($(s0.center) + (5mm, -19mm)$) {};
		\node[mappoint] (mm15) at ($(s0.center) + (2mm, -20mm)$) {};
		\node[mappoint] (mm16) at ($(s0.center) + (-1mm, -18mm)$) {};
		
		% ICP factors
		\draw[graphedge] (s1.south) edge[bend left=20] node[midway, icpfactor] {} (mm1);
		\draw[graphedge] (s2.south) edge[bend left=20] node[midway, icpfactor] {} (mm2);
		\draw[graphedge] (s3.south west) edge[bend left=20] node[midway, icpfactor] {} (mm3);
		\draw[graphedge] (s4.south west) edge[bend left=20] node[midway, icpfactor] {} (mm4);
		\draw[graphedge] (s5.south west) edge[bend left=20] node[midway, icpfactor] {} (mm5);
		\draw[graphedge] (s6.south west) edge[bend left=20] node[midway, icpfactor] {} (mm6);
		\draw[graphedge] (s7.south west) edge[bend left=20] node[midway, icpfactor] {} (mm7);
		\draw[graphedge] (s8.south west) edge[bend left=20] node[midway, icpfactor] {} (mm8);
		
		% legend
		\coordinate (legend) at ($(s0.west)!0.3!(s8.east) + (0mm, -32mm)$);
		\node[state, left=35mm of legend, label={[font=\footnotesize, label distance=0.5mm]right:Estimated State}] (c2) {};
		\node[dot, black, minimum size=1mm, left=12mm of legend, yshift=0mm, label={[font=\footnotesize, label distance=1mm]right:Local Map Point}] {};
		\node[interpolated, left=-14.5mm of legend, yshift=0mm, label={[font=\footnotesize, label distance=0.5mm]right:Interpolated State}] {};
		
		\node[dot, red, left=12mm of legend, yshift=-5mm, label={[font=\footnotesize, label distance=0.5mm]right:Motion Prior $J_v$}] {};
		\node[dot, blue, left=38mm of legend, yshift=-5mm, label={[font=\footnotesize, label distance=0.5mm]right:ICP Factor $J_{\text{p2p}}$}] {};
		\node[dot, green, left=-14mm of legend, yshift=-5mm, label={[font=\footnotesize, label distance=0.5mm]right:IMU Factor $J_{\text{imu}}$}] {};
	\end{tikzpicture}
	\else
	\includegraphics{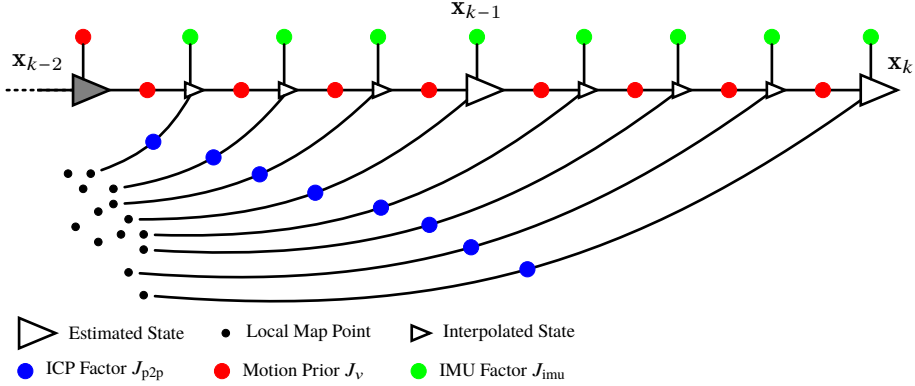}
	\fi
	\caption{This figure depicts a factor graph of our sliding window lidar-inertial odometry using a continuous-time motion prior. The larger triangles represent the estimation times that form our sliding window. The state $\mathbf{x}(t) = \{\mathbf{T}(t), \boldsymbol{\varpi}(t), \dot{\boldsymbol{\varpi}}(t), \mathbf{b}(t) \}$ includes the pose $\mathbf{T}(t)$, the body-centric velocity $\boldsymbol{\varpi}(t)$, the body-centric acceleration $\dot{\boldsymbol{\varpi}}(t)$, and the IMU biases $\mathbf{b}(t)$. The grey-shaded state $\mathbf{x}_{k-2}$ is next to be marginalized and is held fixed during the optimization of the current window. The smaller triangles are interpolated states that we do not directly estimate during the optimization process. Instead, we construct continuous-time measurement factors using the posterior Gaussian process interpolation formula. We include a unary prior on $\mathbf{x}_{k-2}$ to denote the prior information from the sliding window filter}
	\label{fig:cticp_factor_graph}
\end{figure*}

\section{Lidar-Inertial Odometry}

%In our notation, $\boldsymbol{\varpi}_v^{iv}$ is the six degree of freedom body-centric velocity from the vehicle frame to the inertial frame as measured in the vehicle frame.

% TODO: where does sparsity come in (sparse inverse kernel matrix) and how does this affect interpolation

%Due to our choice to model the prior as a sequence of local LTI SDEs with a Markovian state, ... cite Barfoot, at the estimation times, $x \ sim ...$, define the kernel matrix, state that the inverse kernel matrix is exactly sparse so batch trajectory estimation as exactly sparse Gaussian process regression is linear in the number of measurement times. Also, the sparsity of the inverse kernel matrix means that interpolation is an O(1) operation. (Give equations for this.)

% exact sparsity, in trajectory estimation (no landmarks) the inverse kernel matrix exhibits a block-tridiagonal structure which means that Gaussian process regression can be performed in O(m) time. The exact sparsity of the inverse kernel matrix is also what allows us to perform efficient Gaussian process interpolation.

% we refer the reader to X, Y, Z for further details.

Our lidar-inertial odometry is implemented as sliding-window batch trajectory estimation. The factor graph corresponding to our approach is depicted in Figure~\ref{fig:cticp_factor_graph}. The state $\mathbf{x}(t) = \{\mathbf{T}(t), \boldsymbol{\varpi}(t), \dot{\boldsymbol{\varpi}}(t), \mathbf{b}(t)\}$ consists of the $SE(3)$ pose $\mathbf{T}_{vi}(t)$, the body-centric velocity $\boldsymbol{\varpi}_v^{vi}(t)$, the body-centric acceleration $\dot{\boldsymbol{\varpi}}_v^{vi}(t)$, and the IMU biases $\mathbf{b}(t)$. $\boldsymbol{\varpi}_v^{vi}$ is a $6\times1$ vector containing the body-centric linear velocity $\boldsymbol{\nu}_v^{vi}$ and angular velocity $\boldsymbol{\omega}_v^{vi}$. We approximate the $SE(3)$ trajectory using a sequence of local Gaussian processes as in \cite{anderson_iros15}. Between pairs of estimation times, the local variable $\boldsymbol{\xi}_k(t)$ is defined as 
 \begin{equation}
 	\boldsymbol{\xi}_k(t) = \ln(\mathbf{T}(t) \mathbf{T}(t_k)^{-1})^\vee. \label{eq:local_var}
 \end{equation} We use \eqref{eq:local_var} and the following to convert between global and local variables:
 \begin{align}
 	\dot{\boldsymbol{\xi}}_k(t) &= \J(\boldsymbol{\xi}_k(t))^{-1} \boldsymbol{\varpi}(t),  \\
 	\ddot{\boldsymbol{\xi}}_k(t) &\approx -\frac{1}{2}(\J(\x_k(t))^{-1} \w(t))^\curlywedge \w(t) + \J(\x_k(t))^{-1} \dot{\w}(t),
 \end{align} where the approximation for $\ddot{\boldsymbol{\xi}}_k(t)$ was originally derived by Tang et al. \cite{tang_ral19}. We use a Singer prior, introduced in \cite{wong_ral20}, which is defined by the following Gaussian process,
 \begin{equation}
 	\ddot{\x}_k(t) \sim \mathcal{GP}(\mathbf{0}, \boldsymbol{\sigma}^2 \exp(-\boldsymbol{\ell}^{-1}|t - t^\prime|)),
 \end{equation} and which can equivalently be represented using the following linear time-invariant stochastic differential equation,
 \begin{align}
 	\dot{\boldsymbol{\gamma}}_k(t) &= \mathbf{A} \boldsymbol{\gamma}_k(t) + \mathbf{L} \mathbf{w}(t), \label{eq:singer}
 \end{align} \vspace{-2mm} where \begin{align}
 	\mathbf{w}(t) &\sim \mathcal{GP}(\mathbf{0}, \mathbf{Q}_c \delta(t - t^\prime)), \nonumber \\
 	\boldsymbol{\gamma}_k(t) &= \begin{bmatrix} \boldsymbol{\xi}_k(t) \\ \dot{\boldsymbol{\xi}}_k(t) \\ \ddot{\boldsymbol{\xi}}_k(t) \end{bmatrix},~\mathbf{A} = \begin{bmatrix} \mathbf{1} & \mathbf{0} & \mathbf{0} \\ \mathbf{0}& \mathbf{1} & \mathbf{0} \\ \mathbf{0}& \mathbf{0} & -\boldsymbol{\alpha} \end{bmatrix},~\mathbf{L} = \begin{bmatrix} \mathbf{0} \\ \mathbf{0} \\ \mathbf{1} \end{bmatrix}, \nonumber
 \end{align} $\boldsymbol{\sigma}^2$ is a variance, $\boldsymbol{\ell}$ is a length scale, $\boldsymbol{\alpha} = \boldsymbol{\ell}^{-1}$, and $\mathbf{w}(t)$ is a white-noise Gaussian process where $\mathbf{Q}_c = 2\boldsymbol{\alpha}\boldsymbol{\sigma}^2$ is the associated power spectral density matrix. \eqref{eq:singer} can be stochastically integrated to arrive at a local Gaussian process
 \begin{align}
 	\boldsymbol{\gamma}_k(t) \sim \mathcal{GP}(&\boldsymbol{\Phi}(t, t_k) \check{\boldsymbol{\gamma}}_k(t_k)), \boldsymbol{\Phi}(t, t_k) \check{\mathbf{P}}(t_k) \boldsymbol{\Phi}(t, t_k)^T +  \change{\mathbf{Q}_t}   ),
 \end{align} where the formulation for the transition function $\boldsymbol{\Phi}(t, t_k)$ and the covariance \change{$\mathbf{Q}_t$} can be found in \cite{wong_ral20}. In order to convert our continuous-time formulation into a factor graph, we build a sequence of motion prior factors between pairs of estimation times using
 \begin{subequations}
 	\begin{align}
 		J_{v,k} &= \frac{1}{2} \mathbf{e}_{v,k}^T \mathbf{Q}_k^{-1} \mathbf{e}_{v,k}, \label{eq:prior1} \\
 		\mathbf{e}_{v,k} &= \boldsymbol{\gamma}_k(t_{k+1}) - \boldsymbol{\Phi}(t_{k+1}, t_k) \boldsymbol{\gamma}_k(t_k).  \label{eq:prior3}
 	\end{align}
 \end{subequations} Our IMU measurement model is
\begin{equation}
	\begin{bmatrix} \tilde{\mathbf{a}} \\ \tilde{\boldsymbol{\omega}} \end{bmatrix} = \begin{bmatrix}  \mathbf{a}_v^{vi} - \mathbf{C}_{vi}\mathbf{g}_i \\ \boldsymbol{\omega}_v^{vi} \end{bmatrix} + \begin{bmatrix} \mathbf{b}_a \\ \mathbf{b}_{\omega} \end{bmatrix} + \begin{bmatrix} \mathbf{w}_a \\ \mathbf{w}_{\omega} \end{bmatrix}, \label{eq:imu_meas_model}
\end{equation} where $\mathbf{b}_a$ and $\mathbf{b}_\omega$ are the accelerometer and gyroscope biases, $\mathbf{w}_a \sim \mathcal{N}(\mathbf{0}, \mathbf{R}_{a})$ and $\mathbf{w}_\omega \sim \mathcal{N}(\mathbf{0}, \mathbf{R}_{\omega})$ are zero-mean Gaussian noise. Due to angular velocity and acceleration being a part of the state, the associated IMU error function is straightforward:
\begin{subequations} \label{eq:gyro_error}
	\begin{align} 
		J_{\text{imu}, \ell} &= \change{\frac{1}{2} \begin{bmatrix} \mathbf{e}_{a,\ell} \\ \mathbf{e}_{\omega,\ell} \end{bmatrix}^T \begin{bmatrix} \mathbf{R}_a & \\ & \mathbf{R}_\omega \end{bmatrix}^{-1} \begin{bmatrix} \mathbf{e}_{a,\ell} \\ \mathbf{e}_{\omega,\ell} \end{bmatrix},} \\
		\mathbf{e}_{a,\ell} &= \tilde{\mathbf{a}}_\ell - \dot{\boldsymbol{\nu}}(\tau_\ell) + \mathbf{C}_{vi}(\tau_\ell) \mathbf{g}_i -\mathbf{b}_{a}(\tau_\ell), \\
		\mathbf{e}_{\omega,\ell} &= \tilde{\boldsymbol{\omega}}_\ell - \boldsymbol{\omega}(\tau_{\ell}) - \mathbf{b}_{\omega}(\tau_{\ell}),
	\end{align}
\end{subequations} where we rely on forming measurement factors using the posterior Gaussian process interpolation formula. For each of these continuous-time measurement factors, we compute Jacobians of the perturbation to the state at the interpolated times with respect to the bracketing estimation times. This is an approximation, as it is not exactly the same as marginalization. However, we have found it to work well in practice. We also include motion prior factors for the IMU biases,
\begin{subequations}
	\begin{align}J_{v,b,k} &= \frac{1}{2} \mathbf{e}_{v,b,k}^T \mathbf{Q}_{b,k}^{-1} \mathbf{e}_{v,b,k}, \\
		\mathbf{e}_{v,b,k} &= \mathbf{b}(t_{k+1}) - \mathbf{b}(t_{k}),
	\end{align}
\end{subequations} where $\mathbf{Q}_{b,k} = \bm{Q}_b \Delta t_k$ is the covariance resulting from a white-noise-on-velocity motion prior, and $\bm{Q}_b$ is the associated power spectral density matrix. We use point-to-plane factors. The associated error function is
\begin{subequations}
	\begin{align}
		J_{\text{p2p},j} &= \mathbf{e}_{\text{p2p},j}^T \mathbf{R}_{\text{p2p},j}^{-1} \mathbf{e}_{\text{p2p},j}, \\
		\mathbf{e}_{\text{p2p},j} &= \change{\alpha_j} \mathbf{n}_j^T \mathbf{D} (\mathbf{p}_j - \mathbf{T}_{vi}(\tau_j)^{-1} \mathbf{T}_{vs} \mathbf{q}_j),
	\end{align}
\end{subequations} where $\mathbf{q}_j$ is the query point, $\mathbf{p}_j$ is the matched point in the local map, $\mathbf{n}_j$ is an estimate of the surface normal at $\mathbf{p}_j$ given neighboring points in the map, $\mathbf{D}$ is a constant matrix removing the homogeneous component, $\mathbf{T}_{vs}$ is a known extrinsic calibration between the lidar frame and the vehicle frame, and $\change{\alpha_j} = (\sigma_2 - \sigma_3)/\sigma_1$ \cite{dellenbach_icra22} is a heuristic weight to favour planar neighborhoods. The objective function that we seek to minimize is
\begin{equation}
	J = \sum_k J_{v,k} + \sum_j J_{\text{p2p},j} + \sum_\ell J_{\text{imu},\ell}.
\end{equation}
We solve this nonlinear least squares problem for the optimal perturbation to the state using Gauss-Newton. Once the solver has converged, we update the pointcloud correspondences and iterate this two-step process to convergence. In practice, we typically limit the maximum number of inner-loop Gauss-Newton iterations to \change{10}, and the number of outer-loop iterations to 10 in order to enable real-time operation.

In our approach, we estimate the orientation of the gravity vector relative to the initial map frame at startup. We perform sliding-window batch trajectory \change{estimation} where the length of the sliding window is $200$ms. We output the pose at the middle of the newest lidar scan.

\section{IMU-as-Input Lidar-Inertial Baseline}

\begin{figure*}[t]
	\centering
	\ifpreprint
	\begin{tikzpicture}[
		line cap=round,
		graphedge/.style={>=latex, line width=1pt},
		state/.style={draw, thick, line width=1pt, isosceles triangle,isosceles triangle apex angle=45, minimum size=4mm, inner sep=0pt, outer sep=0pt, fill=white},
		interpolated/.style={state, minimum size=2mm},
		dot/.style={draw, inner sep=0, circle, fill, minimum size=2mm, line width=0pt},
		mappoint/.style={dot, black, minimum size=1mm, outer sep=1mm},
		vfactor/.style={dot, green},
		preintfactor/.style={dot, magenta},
		pfactor/.style={dot, red},
		icpfactor/.style={dot, blue},
		every label/.style={align=left}
		]
		
		% main vertices and edges
		\node[] (p1) {};
		\node[] (p2) [right= of p1.center] {};
		% initial link
		\draw[graphedge, dotted] (p1) -- (p2);
		\begin{scope}
			\clip (p2) circle (6mm);
			\draw[graphedge] (p1) -- (p2);
		\end{scope}
		
		\def \dx{0mm};
		\def \vy{0.7};
		% state triangles
		\node[state, fill=gray, label=above left:{$\mathbf{x}_{k-2}$}] (s0) [right = of p1.center] {};
		\node[interpolated] (s1) [right =of s0, xshift=\dx] {};
		\node[interpolated] (s2) [right =of s1, xshift=\dx] {};
		\node[state,  label=above right:{$\mathbf{x}_{k-1}$}] (s3) [right =of s2, xshift=\dx] {};
		\node[interpolated] (s4) [right =of s3, xshift=\dx] {};
		\node[interpolated] (s5) [right =of s4, xshift=\dx] {};
		\node[state, label=above right:{$\mathbf{x}_{k}$}] (s6) [right =of s5, xshift=\dx] {};
		\node[interpolated] (s7) [right =of s6, xshift=\dx] {};
		\node[interpolated] (s8) [right =of s7, xshift=\dx] {};

		% Prior factors
		\draw[graphedge] (s0.east) edge node[midway] (s01) {} (s1.west);
		\draw[graphedge] (s1.east) edge node[midway] (s12) {} (s2.west);
		\draw[graphedge] (s2.east) edge (s3.west);
		\draw[graphedge] (s3.east) edge (s4.west);
		\draw[graphedge] (s4.east) edge node[midway] (s45) {} (s5.west);
		\draw[graphedge] (s5.east) edge (s6.west);
		\draw[graphedge] (s6.east) edge (s7.west);
		\draw[graphedge] (s7.east) edge (s8.west);
		
		% sliding window prior
		\node[pfactor] (v0) at ($(s0.center) + (0, \vy) $) {};
		
		% velocity factor
%		\node[vfactor] (v1) at ($(s1.center) + (0, \vy)$) {};
%		\node[vfactor] (v2) at ($(s2.center) + (0, \vy)$) {};
%		\node[vfactor] (v3) at ($(s3.center) + (0, \vy)$) {};
%		\node[vfactor] (v4) at ($(s4.center) + (0, \vy)$) {};
%		\node[label={$\mathbf{x}_{k-1}$}]  (ttt) at ($(s4.center) + (0, \vy)$) {};
%		\node[vfactor] (v5) at ($(s5.center) + (0, \vy)$) {};
%		\node[vfactor] (v6) at ($(s6.center) + (0, \vy)$) {};
%		\node[vfactor] (v7) at ($(s7.center) + (0, \vy)$) {};
%		\node[vfactor] (v8) at ($(s8.center) + (0, \vy)$) {};
		
		\draw[graphedge] (s0) -- (v0);
%		\draw[graphedge] (s1) -- (v1);
%		\draw[graphedge] (s2) -- (v2);
%		\draw[graphedge] (s3) -- (v3);
%		\draw[graphedge] (s4) -- (v4);
%		\draw[graphedge] (s5) -- (v5);
%		\draw[graphedge] (s6) -- (v6);
%		\draw[graphedge] (s7) -- (v7);
%		\draw[graphedge] (s8) -- (v8);
		
		% local map points
		\node[mappoint] (mm1) at ($(s0.center) + (1mm, -11mm)$) {};
		\node[mappoint] (mm2) at ($(s0.center) + (4mm, -13mm)$) {};
		\node[mappoint] (mm3) at ($(s0.center) + (4mm, -15mm)$) {};
		\node[mappoint] (mm4) at ($(s0.center) + (6mm, -17mm)$) {};
		\node[mappoint] (mm5) at ($(s0.center) + (8mm, -19mm)$) {};
		\node[mappoint] (mm6) at ($(s0.center) + (8mm, -21mm)$) {};
		\node[mappoint] (mm7) at ($(s0.center) + (6mm, -24mm)$) {};
		\node[mappoint] (mm8) at ($(s0.center) + (8mm, -27mm)$) {};
		\node[mappoint] (mm11) at ($(s0.center) + (-2mm, -11mm)$) {};
		\node[mappoint] (mm12) at ($(s0.center) + (0mm, -13mm)$) {};
		\node[mappoint] (mm13) at ($(s0.center) + (2mm, -16mm)$) {};
		\node[mappoint] (mm14) at ($(s0.center) + (5mm, -19mm)$) {};
		\node[mappoint] (mm15) at ($(s0.center) + (2mm, -20mm)$) {};
		\node[mappoint] (mm16) at ($(s0.center) + (-1mm, -18mm)$) {};
		
		% ICP factors
		\draw[graphedge] (s1.south) edge[bend left=20] node[midway, icpfactor] {} (mm1);
		\draw[graphedge] (s2.south) edge[bend left=20] node[midway, icpfactor] {} (mm2);
		\draw[graphedge] (s3.south west) edge[bend left=20] node[midway, icpfactor] {} (mm3);
		\draw[graphedge] (s4.south west) edge[bend left=20] node[midway, icpfactor] {} (mm4);
		\draw[graphedge] (s5.south west) edge[bend left=20] node[midway, icpfactor] {} (mm5);
		\draw[graphedge] (s6.south west) edge[bend left=20] node[midway, icpfactor] {} (mm6);
		\draw[graphedge] (s7.south west) edge[bend left=20] node[midway, icpfactor] {} (mm7);
		\draw[graphedge] (s8.south west) edge[bend left=20] node[midway, icpfactor] {} (mm8);
		
		% preintegrated velocity factor
		\node[pfactor] (pi2) at ($(s12.center) + (0, \vy) + (0, \vy)$) {};
		\node[pfactor] (pi6) at ($(s45.center) + (0, \vy) + (0, \vy)$) {};
		\draw[graphedge] (s0.north) edge[bend left=20] (pi2);
		\draw[graphedge] (pi2) edge[bend left=20] (s3.north);
		\draw[graphedge] (s3.north) edge[bend left=20] (pi6);
		\draw[graphedge] (pi6) edge[bend left=20] (s6.north);
		
		% legend
		\coordinate (legend) at ($(s0.west)!0.3!(s8.east) + (0mm, -32mm)$);
		\node[state, left=35mm of legend, label={[font=\footnotesize, label distance=0.5mm]right:Estimated State}] (c2) {};
		\node[dot, black, minimum size=1mm, left=12mm of legend, yshift=0mm, label={[font=\footnotesize, label distance=1mm]right:Local Map Point}] {};
		\node[interpolated, left=-14.5mm of legend, yshift=0mm, label={[font=\footnotesize, label distance=0.5mm]right:Extrapolated State (Using IMU Integration)}] {};
		
		\node[dot, red, left=12mm of legend, yshift=-5mm, label={[font=\footnotesize, label distance=0.5mm]right:Classic Preintegration Factor $J_v$}] {};
		\node[dot, blue, left=38mm of legend, yshift=-5mm, label={[font=\footnotesize, label distance=0.5mm]right:ICP Factor $J_{\text{p2p}}$}] {};
%		\node[dot, green, left=-14mm of legend, yshift=-5mm, label={[font=\footnotesize, label distance=0.5mm]right:IMU Factor $J_{\text{imu}}$}] {};
	\end{tikzpicture}
	\else
	\includegraphics{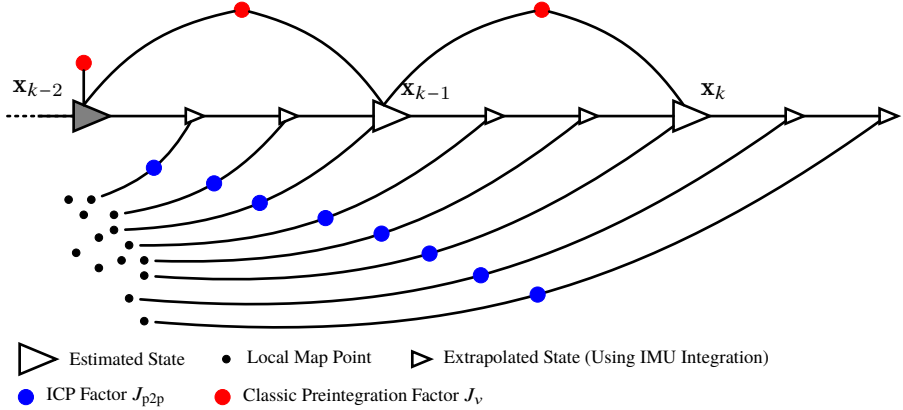}
	\fi
	\caption{\change{This figure depicts a factor graph of our baseline approach that uses IMU measurements to de-skew the pointcloud and to form relative motion priors using classic preintegration. The larger triangles represent the estimation times that form our sliding window. The state $\mathbf{x}(t_k) = \{\mathbf{C}_{iv}(t_k), \mathbf{r}_i^{vi}(t_k), \mathbf{v}^{vi}_i(t_k), \mathbf{b}(t_k) \}$ includes the orientation and position in a global frame, the velocity in a global frame, and the IMU biases. The grey-shaded state $\mathbf{x}_{k-2}$ is next to be marginalized. The smaller triangles are extrapolated states that we do not directly estimate during the optimization process. Instead, we extrapolate for these states using IMU integration starting at an estimated state. The factor graph includes a unary prior on $\mathbf{x}_{k-2}$ to denote the prior information from the sliding window filter}}
	\label{fig:baseline_factor_graph}
\end{figure*}

\change{
As a baseline where IMU measurements are treated as an input, we consider a lidar-inertial odometry approach where IMU measurements are used to de-skew the lidar pointcloud and classic preintegration is used as a prior. The baseline is implemented as sliding-window batch trajectory estimation and the factor graph corresponding to the baseline approach is depicted in Figure~\ref{fig:baseline_factor_graph}. The state $\mathbf{x}(t_k) = \{\mathbf{C}_{iv}(t_k), \mathbf{r}_i^{vi}(t_k), \mathbf{v}^{vi}_i(t_k), \mathbf{b}(t_k) \}$ consists of the orientation $\mathbf{C}_{iv}(t_k)$, position $\mathbf{r}_i^{vi}(t_k)$, velocity $\mathbf{v}^{vi}_i(t_k)$, and IMU biases. All variables are expressed in a global frame. We use classic preintegration to form binary factors between pairs of estimated states in the sliding window \cite{forster_tro17}. At each iteration of the optimization, we integrate the IMU measurements to extrapolate for the state at each IMU measurement time using \begin{subequations}
	\begin{align}
		\mathbf{C}_j &= \mathbf{C}_i \prod_{k=i}^{j-1} \exp\left(\Delta t_k  (\tilde{\boldsymbol{\omega}}_k - \mathbf{b}_\omega(t_k))^\wedge \right ),  \\
		\mathbf{v}_j &= \mathbf{v}_i + \mathbf{g} \Delta t_{ij} + \sum_{k=i}^{j-1} \mathbf{C}_k (\tilde{\mathbf{a}}_k - \mathbf{b}_a(t_k)) \Delta t_k, \\
		\mathbf{r}_j &= \mathbf{r}_i + \sum_{k=i}^{j-1} \left[ \mathbf{v}_k \Delta t_k + \frac{1}{2} \mathbf{g} \Delta t_k^2 + \frac{1}{2}\mathbf{C}_k (\tilde{\mathbf{a}}_k - \mathbf{b}_a(t_k)) \Delta t_k^2 \right].
	\end{align}
\end{subequations} The position at a given lidar point time can then be obtained by linearly interpolating between the positions at the IMU measurement times. The orientation at a given lidar point time can be obtained using the following formula, \begin{equation}
	\mathbf{C}(\tau_j) = \mathbf{C}(t_\ell) \left(\mathbf{C}(t_\ell)^T \mathbf{C}(t_{\ell+1}) \right)^\alpha
\end{equation} where $\alpha = (\tau_j - t_\ell) / (t_{\ell+1} - t_\ell)$. Using these interpolated states, we can write the point-to-plane error function as \begin{equation} \label{eq:baseline_error}
		\mathbf{e}_{\text{p2p},j} = \mathbf{n}_j^T \left(\mathbf{p}_j - \mathbf{C}_{iv}(\tau_j)(\mathbf{C}_{vs}\mathbf{q}_j +\mathbf{r}_v^{sv}) - \mathbf{r}_i^{vi}(\tau_j) \right).
\end{equation} The Jacobians of this error function with respect to perturbations to the state variables are provided in the appendix.

}

\section{Lidar-Inertial Simulation}

\begin{figure}[t]
	\centering
	\includegraphics[width=0.5\columnwidth]{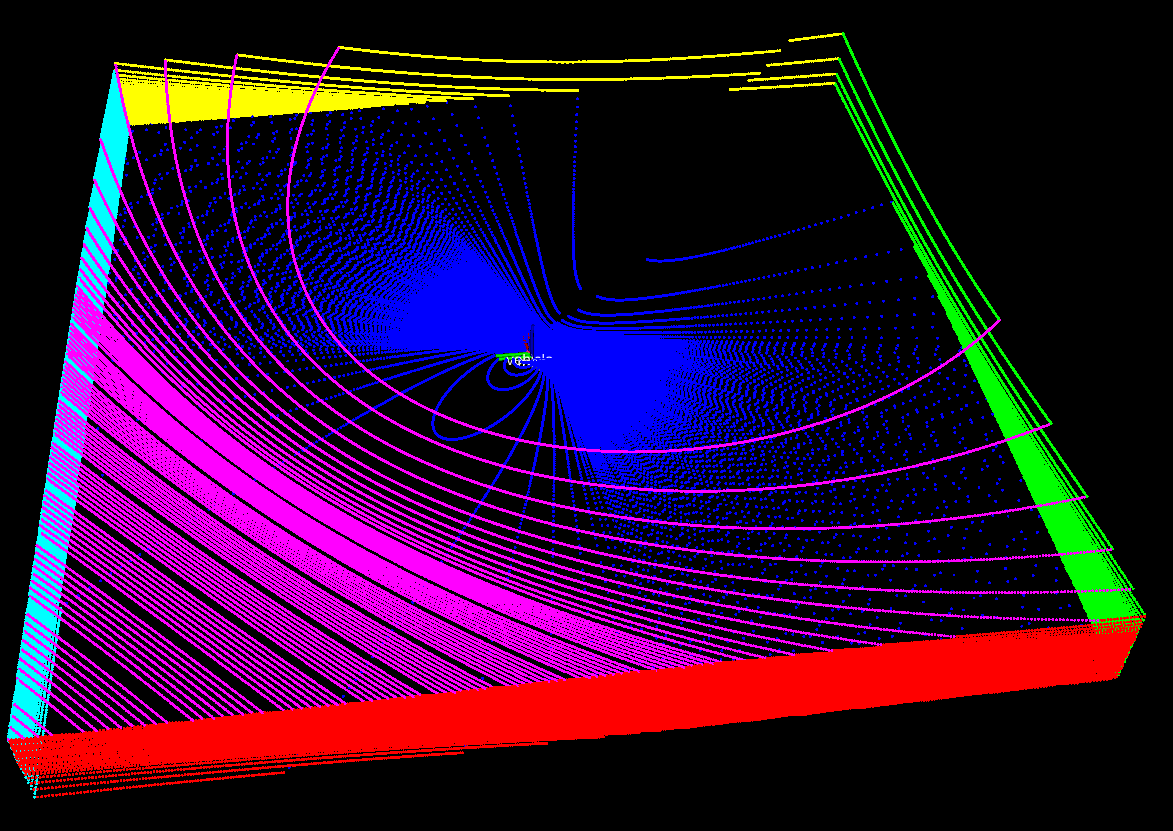}
	\caption{This figure depicts an example lidar pointcloud produced by our simulation, which contains motion distortion. The pointcloud is colored based off which wall the lidar point is reflected}
	\label{fig:sim_lidar}
\end{figure}

\begin{figure}[t]
	\centering
	\ifpreprint
	\includegraphics[width=0.6\columnwidth]{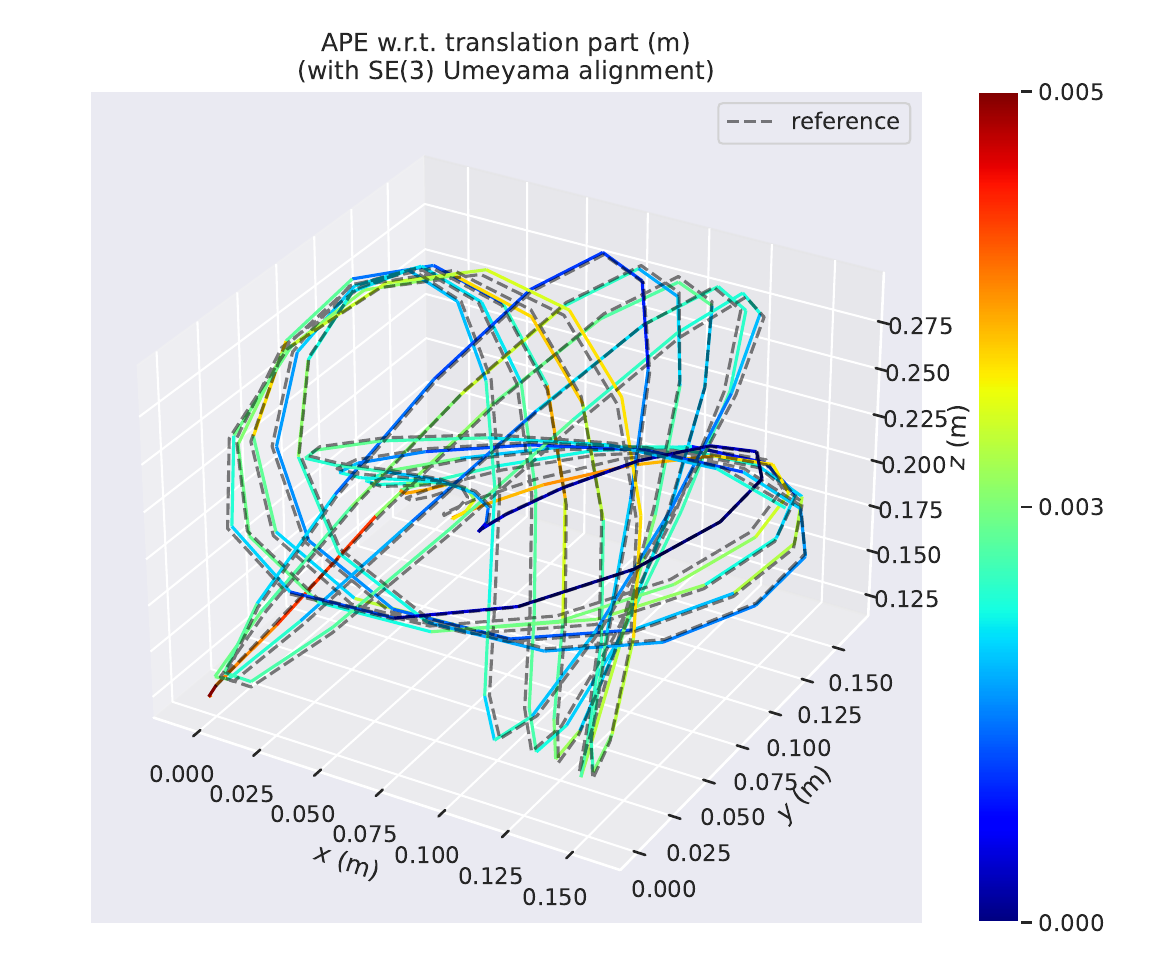}
	\else
	\includegraphics[width=0.6\columnwidth]{epsfigs/slow.eps}
	\fi
	\caption{\change{This figure depicts the results of our lidar-inertial simulation where the ground truth position (dashed line) is compared to the position estimated by Singer-LIO colored by the absolute position error. This trajectory is an example of one of the slow sequences}}
	\label{fig:sim_results}
\end{figure}

\change{In this section, we compare the performance of our lidar-inertial odometry to the baseline imu-as-input approach in a simulated environment. The simulated environment is a rectangular room and we simulate trajectories using sinusoidal body-centric velocities,}
\begin{equation}
	\left[\w(t)\right]_j = A_j \sin(2\pi f_j t),
\end{equation} where $A_j$ and $f_j$ are configurable amplitude and frequency parameters. The resulting body-centric acceleration can be obtained via differentiation,
\begin{equation}
	\left[\dot{\w}(t)\right]_j = A_j 2\pi f_j \cos(2\pi f_j t).
\end{equation} We then step through the simulation so as to replicate the lidar firing sequence of a Velodyne Alpha-Prime 128-beam lidar, obtaining the pose of the sensor for each firing sequence. Starting with $\mathbf{T}_0 = \mathbf{T}_{vi}(t_0) = \mathbf{1}$,
\begin{equation}
	\mathbf{T}_{k+1} \approx \exp\left( \left(\w(t_k)\Delta t_k + \frac{1}{2} \Delta t_k^2 \dot{\w}(t_k)\right)^\wedge \right) \mathbf{T}_{k},
\end{equation} \change{where $\Delta t_k$ is very small (53.3$\mu$s). By generating the trajectories in this way, it is straightforward to extract the body-centric angular velocity and linear acceleration to simulate IMU measurements. We use the measurement model in \eqref{eq:imu_meas_model} to simulate biases and gravity components. \change{We simulate a constant nonzero bias on each gyroscope and accelerometer axis.} We include zero-mean Gaussian noise on IMU measurements as well as Gaussian noise on the range measurement of each lidar point. We chose measurement noises to be close to what we experience on our experimental platform. Figure~\ref{fig:sim_lidar} depicts an example pointcloud produced in our simulation environment where the points are colored based off which wall they are reflected. Figure~\ref{fig:sim_results} compares the trajectory estimated by our lidar-inertial odometry with the ground truth.}

\change{For the simulation parameters, we use an IMU rate of 200Hz, a simulation length of 20s, and three motion regimes denoted \textit{slow}, \textit{medium}, and \textit{fast}. Where for each of these motion regimes, we randomly sample for the amplitudes and frequencies of the body-centric velocities used in the simulation of a sequence. The ranges for these parameters is given in Table~\ref{tab:sim-parameters}. One set of amplitudes and frequencies is sampled for each of the 20 sequences simulated for the three motion regimes.We set the standard deviation of the accelerometer measurement noise to 0.02 $\text{m/s}^2$, the standard deviation of the gyroscope measurement noise to 0.01 rad/s, and the standard deviation of the lidar range measurements to 0.02m. The accelerometers were given a constant bias of 0.05  $\text{m/s}^2$ and the gyroscopes were given a constant bias of 0.05 rad/s.}

\begin{table}[t]
	\centering
	\caption{\change{Simulation parameter ranges for the different motion regimes}}
	\label{tab:sim-parameters}
	\begin{tabular}{ l ? c c c }
		\toprule
		& Slow & Medium & Fast \\
		\midrule
Linear Velocity Amplitude [m/s]  &   $A \in [0.1, 0.5]$   &  $A \in [0.5, 1.0]$      &  $A \in [1.0, 2.0]$   \\
Angular Velocity Amplitude [rad/s] &  $A \in [0.1, 0.5]$    &   $A \in [0.5, 1.0]$     &   $A \in [1.0, 2.0]$    \\
Linear Velocity Frequency [Hz]  &  $f \in [0.5, 1.0]$    & $f \in [1.0, 2.0]$        &   $f \in [2.0, 4.0]$   \\
Angular Velocity Frequency [Hz] & $f \in [1.0, 2.0]$     &  $f \in [2.0, 4.0]$      &  $f \in [4.0, 8.0]$   \\
\bottomrule
\end{tabular}
\end{table}

%\begin{align}
%	A &= \{1.0, 1.0, 1.0, 0.5, 0.5, 0.5\}, \nonumber \\
%	f &= \change{\{1.0, 1.0, 1.0, 1.0, 1.0, 1.0\}}, \nonumber \\
%	\mathbf{b} &= [0.01, 0.01, 0.01, 0.05, 0.05, 0.05]^T, \nonumber \\
%	\mathbf{C}_{ig} &= \exp(\boldsymbol{\phi}_{ig}^\wedge), \boldsymbol{\phi}_{ig} = [-0.0197052, 0.0285345, 0.]^T, \nonumber\\
%	\mathbf{R}_a^{1/2} &= \text{diag}(0.01249, 0.01249, 0.01249), \nonumber \\
%	\mathbf{R}_\omega^{1/2} &= \text{diag}(0.00208, 0.00208, 0.00208), \nonumber \\
%	\sigma_{\text{lidar}} &= 0.02. \nonumber
%\end{align} 
\change{We compare the performance of our lidar-inertial odometry against the baseline in Table~\ref{tab:sim-quantitative} where we also show the performance of our approach using only the lidar and the gyroscope, and lidar only. We obtained the results by computing the absolute trajectory error between our estimated trajectories and the ground truth using the evo evaluation tool\footnote{\url{https://github.com/MichaelGrupp/evo}}. The results in the table are the overall root mean squared error obtained by concatenating the error from each individual sequence. The results show that in the low speed regime, the imu-as-input baseline approach and our imu-as-measurement approach based on the Singer prior achieve nearly identical results. This is unsurprising as it appears to replicate the results from Section~\ref{section:sim_results}. Interestingly, our lidar-only approach performs the best on the slow regime. However, in the medium and fast regime, the advantage of our lidar-inertial approach becomes apparent. In the medium regime, the baseline imu-as-input approach begins to break down. This is possibly due to the fact that the motion is no longer approximately constant acceleration and constant angular rate. On the other hand, our lidar-inertial approach performs roughly the same in the medium regime. Our lidar-only approach also breaks down in the medium regime. In the fast regime, both our lidar-inertial and lidar with gyro approaches achieve respectable results while the lidar-only approach and the imu-as-input baseline fail completely.
}

%From the results, we can see that the addition of IMU measurements improves performance measurably.

%$A_j = \{1.0, 1.0, 1.0, 0.5, 0.5, 0.5\}$, $f_j = \{1.0, 1.0, 1.0, 1.0, 1.0, 1.0, 1.0\}$, $\mathbf{b} = \{0.01, 0.01, 0.01, 0.05, 0.05, 0.05\}$, $\mathbf{C}_{ig} = \exp(\boldsymbol{\phi}_{ig}^\wedge), \boldsymbol{\phi}_{ig} = [-0.0197052, 0.0285345, 0.]^T$, $\mathbf{R}_a^{1/2} = \text{diag}(0.0195, 0.0195, 0.0195)$, $\mathbf{R}_\omega^{1/2} = \text{diag}(0.01, 0.01, 0.01)$, $\sigma_{\text{lidar}} = 0.02m$, IMU rate is 200 Hz.

%\begin{table}[t]
%	\centering
%	\caption{Quantitative results in our lidar-inertial simulation.}
%	\label{tab:sim-quantitative}
%	\begin{tabular}{ l ? c c c }
%		\toprule
%		\midrule
%		\textbf{Simulation}              & STEAM-LO & STEAM-LO + Gyro & STEAM-LIO \\
%		\midrule
%		max & 0.009957 & 0.007493 & 0.007518\\
%		mean & 0.004147 & 0.002223 & 0.002234\\
%		median & 0.004109 & 0.001962 & 0.002016 \\
%		min & 0.000577 & 0.000218 & 0.000262\\
%		rmse & 0.004461 & 0.002528 & 0.002542 \\
%		sse & 0.003980 & 0.001278 & 0.001292 \\
%		std & 0.001644 & 0.001204 & 0.001214 \\
%		\midrule
%		\bottomrule
%	\end{tabular}
%\end{table}

%\begin{table}[t]
%	\centering
%	\caption{Quantitative results in our lidar-inertial simulation.}
%	\label{tab:sim-quantitative}
%	\begin{tabular}{ l ? c c c }
%		\toprule
%		\midrule
%		\textbf{Simulation}              & STEAM-LO & STEAM-LO + Gyro & STEAM-LIO \\
%		\midrule
%max & 0.010135 & 0.009440 & 0.012294 \\
%mean & 0.004179 & 0.002514 & 0.002011 \\
%median & 0.004050 & 0.002319 & 0.001699 \\
%min & 0.000615 & 0.000303 & 0.000174 \\
%rmse & 0.004509 & 0.002842 & 0.002493 \\
%sse & 0.004065 & 0.001616 & 0.001243 \\
%std & 0.001693 & 0.001325 & 0.001472 \\
%		\midrule
%		\bottomrule
%	\end{tabular}
%\end{table}

\begin{table}[t]
	\centering
	\caption{\change{Simulation results. Root Mean Squared Absolute Trajectory Error (m). For each speed category (slow, medium, fast), 20 randomized sequences were created. The results in this table are the overall root mean squared absolute position error across 20 sequences, for each approach}}
	\label{tab:sim-quantitative}
	\begin{tabular}{ l ? c c c }
		\toprule
                        & Slow   & Medium & Fast   \\
                        \midrule
			Baseline (IMU as Input) & 0.0026 & 2.1734 & Failed \\
			Singer-LIO              &    0.0026    &    \textbf{0.0025}   &    \textbf{0.0208}     \\
			Singer-LO + Gyro        & 0.0052  & 0.0085 & 0.0445 \\
			Singer-LO               & \textbf{0.0012}      &  12.34      &  Failed     \\
		\bottomrule
	\end{tabular}
\end{table}

%\begin{table}[t]
%	\centering
%	\caption{Quantitative results in our lidar-inertial simulation}
%	\label{tab:sim-quantitative}
%	\begin{tabular}{ l ? c c c }
%		\toprule
%		\textbf{Simulation}              & Singer Prior & Singer Prior + Gyro Only & Singer Prior + IMU \\
%		\midrule
%max	& 0.008520 & 0.004717 & 0.004441 \\ 
%mean	& 0.003913 & 0.000674 & 0.000672 \\
%median	& 0.003931 & 0.000531 & 0.000535 \\
%min	& 0.000787 & 0.000106 & 0.000095 \\
%rmse	& 0.004142 & 0.000912 & 0.000899 \\
%sse	& 0.003430 & 0.000166 & 0.000162 \\
%std	& 0.001357 & 0.000615 & 0.000597 \\
%		\bottomrule
%	\end{tabular}
%\end{table}

%\begin{table}[t]
%	\centering
%	\caption{Quantitative results on Newer College Dataset - Sequence 06, First 950 frames}
%	\label{tab:ncd_06}
%	\begin{tabular}{ l ? c c c }
%		\toprule
%		\textbf{NCD-06}              & Singer Prior & Singer Prior + Gyro Only & Singer Prior + IMU \\
%		\midrule
%max	& 0.298433 & 0.304409 & 0.291731 \\
%mean	& 0.073940 & 0.076383 & 0.076056 \\
%median	& 0.063537 & 0.065579 & 0.064655 \\
%min	& 0.002385 & 0.004366 & 0.007615 \\
%rmse	& 0.084697 & 0.087622 & 0.087065 \\
%sse	& 6.779070 & 7.255313 & 7.163439 \\
%std	& 0.041310 & 0.042933 & 0.042378 \\
%		\bottomrule
%	\end{tabular}
%\end{table}

\section{Experimental Results}

%These types of trajectories are difficult to achieve when working with heavy ground robots but serve as useful tests for our approach.

%It was challenging to obtain results for our lidar-only approach on this dataset. We postulate the reason for this is that our current implementation of lidar-only odometry using the Singer prior struggles with highly dynamic rotations.

%When a highly dynamic rotation is encountered, our approach loses track and the resulting absolute trajectory error (ATE) metrics are too large to be considered useful for comparison. 

%	For a brief segment of sequence 06, we were able to compare lidar-only, lidar + gyro, and lidar-inertial odometry. The results of this comparison are shown in Table~\ref{tab:newer_college} where metrics are computed for the absolute trajectory error. What we observe is that when our lidar-only odometry is working, the performance is quite good. However, after roughly 950 frames, lidar-only odometry fails. When including a gyroscope, we observe much more reliable performance across the dataset. Interestingly, lidar-only odometry actually performs the best on this segment of sequence 06. 

\change{In this section, we provide experimental results on the Newer College Dataset \cite{ramezani_iros20}. This dataset features a 64-beam Ouster lidar and provides the internal IMU measurements of the Ouster lidar. Ground-truth poses were obtained by matching live lidar poses to a map of the environment created using a survey-grade lidar at several stationary poses. This dataset is somewhat unique in that it features several sequences with highly dynamic motions. In Table~\ref{tab:newer_college}, we compare the performance of using our continuous-time Singer prior using lidar only (Singer-LO), lidar and a gyroscope only (Singer-LO + Gyro), and a full lidar-inertial setup including an accelerometer (Singer-LIO). We also compare the performance of our approach to some comparable works in the literature such as CT-ICP \cite{dellenbach_icra22}, a lidar-only approach, FAST-LIO2 \cite{xu_tro22} and DLIO \cite{chen_icra23}, which can be considered the prior state of the art for this dataset, and SLICT \cite{nguyen_ral23} and CLIO \cite{lv_tmech23}, which are two continuous-time approaches that use linear interpolation and B-splines, respectively.
	
Our approach, Singer-LIO, demonstrates the best performance on the 01-Short and 02-Long sequences and also demonstrates the best overall performance. Interestingly, the sequences in which we expected the IMU to make the most difference were 05-Quad w/ Dynamics and 06-Dynamic Spinning due to their dynamic motions. However, we observe that in these sequences, our lidar-only approach performs similarly or even better, replicating the results of our lidar-inertial simulation. It appears that, in this dataset, the addition of an IMU mainly helps in areas where there are geometric degeneracies rather than the areas with dynamic motions. Sequences 05-Quad w/ Dynamics and 06-Dynamic Spinning are very similar to our lidar-inertial simulation in the slow regime, as they are conducted in a rectangular quad at New College, Oxford. FAST-LIO2 and DLIO can be considered state of the art IMU-as-input approaches, and our approach demonstrates a clear advantage over these methods.
}

\begin{table}[t]
	\centering
	\caption{\change{Newer College dataset results.  Root Mean Squared Absolute Trajectory Error (m). Estimated trajectory aligned with ground truth using Umeyama algorithm. $^*$ uses loop closures, $^\dagger$ results obtained from \cite{chen_icra23}, $^\ddagger$ uses camera images}}
	\label{tab:newer_college}
	\resizebox{\textwidth}{!}{\begin{tabular}{ l ? c c c c c ? c}
		\toprule
		\textbf{Newer College Dataset}              & 01-Short            & 02-Long            & 05-Quad w/ Dynamics            & 06-Dynamic Spinning       & 07-Parkland Mound  & \change{\textbf{Overall}}           \\
		\midrule
		CT-ICP$^*$ \cite{dellenbach_icra22} & 0.36          &           &          &              &    &  \\
		\change{KISS-ICP$^\dagger$} \cite{vizzo_ral23} & \change{0.6675} & \change{1.5311} & \change{0.1040} & \change{Failed} & \change{0.2027}  & \\
		FAST-LIO2$^\dagger$ \cite{xu_tro22} & 0.3775 & 0.3324 & 0.0879 & 0.0771 & 0.1483 & \change{0.3152} \\
		DLIO \cite{chen_icra23} & 0.3606 & 0.3268 & \textbf{0.0837} & \textbf{0.0612} & \textbf{0.1196} & \change{0.3048} \\
		SLICT$^*$ \cite{nguyen_ral23} & 0.3843 & 0.3496 & 0.1155 & 0.0844 & 0.1290 & \change{0.3263} \\
		CLIO$^{*\ddagger}$ \cite{lv_tmech23} & 0.408 & 0.381 &  & 0.091 &  & \\
		\midrule
		\change{Singer-LO (Ours)}& \change{0.4543} & \change{Failed} & \change{0.1120} & \change{0.0804} & \change{Failed} & \\
		\change{Singer-LO + Gyro (Ours)} & \change{0.3044} & \change{0.3267} & \change{0.1092} & \change{0.0818} & \change{0.1457} & \change{0.2887} \\
		\change{Singer-LIO (Ours)} & \change{\textbf{0.3020}} & \change{\textbf{0.3186}} & \change{0.1091} & \change{0.0821} & \change{0.1411} &  \change{\textbf{0.2832}} \\
		\bottomrule
	\end{tabular}}
\end{table}

\section{Conclusions}

\change{
In this work, we compared treating an IMU as an input to a motion model vs. treating it as a measurement of the state. On a 1D simulation problem, we showed that these two approaches performed identically when the data is sampled from either a constant velocity or constant acceleration prior and both methods are trained on a hold-out set. We demonstrated our approach to continuous-time lidar-inertial odometry using the Singer prior where body-centric acceleration is included in the state. In our simulated environment, we showed that our lidar-inertial odometry outperformed lidar-only odometry and an IMU-as-input baseline approach. On the Newer College Dataset, we demonstrated state of the art resuts. There is still plenty of work to be done in treating IMU measurements as measurements of the state. Similar to non-uniform B-splines, it would be interesting to investigate a setup where the parameters of the Singer prior are adjusted on the fly so as to adjust between periods of smooth vs. highly dynamic motion. When the IMU is treated as a measurement of the state, this allows us to now incorporate exogenous control inputs into our Gaussian process motion prior. This could be a promising area of research for estimating the state of drones where the torque commanded to the motors is often known. Our approach to combine multiple asynchronous high-rate sensors may prove beneficial in other sensor configurations such as multiple asynchronous IMUs. 
}

%Clearly, there is still room for futher investigation into treating an IMU as a measurement of the state. The experimental results were somewhat surprising, and so clearly further work is needed to get our approach of treating an IMU as a measurement of the state to work reliably. One potential area of concern is that the Gaussian process motion prior does not enforce that the estimated velocity be exactly the integration of the acceleration, but rather it penalizes the state from deviating from this prior. For lidar-only odometry, it may be challenging to include acceleration in the state. In our other work, we investigated combining an IMU with a white-noise-on-acceleration motion prior where the gyroscope is still treated as a measurement of the state but where accelerometer measurements are preintegrated to form relative body-centric velocity factors \cite{burnett_arxiv24}. When using a white-noise-on-acceleration prior, we observed that lidar-only odometry was much more robust than when using a white-noise-on-jerk or Singer prior. It was also comparatively easier to demonstrate that lidar-inertial odometry performed better than lidar-only odometry.

%\ack{} Acknowledgments to be included here.

\ifpreprint

%\newpage
\appendix

\section{Preintegration Using a Schur Complement}

The method of preintegration presented in section~\ref{sec:imu_input} is mathematically equivalent to marginalizing out the unwanted states from the full Bayesian posterior. Marginalization can also be performed using a Schur complement. We will use the Schur complement to efficiently marginalize out unwanted states from our continuous-time formulation. By exploiting sparsity, this can be performed in $O(K)$ time, which is the same time complexity as the classic approach presented in section~\ref{sec:imu_input}.

We consider the factor graph shown in Figure~\ref{fig:factor}, which could potentially be a result of our continuous-time state estimation with binary motion prior factors, unary measurement factors, and a unary prior factor on the initial state $\mathbf{x}_0$. Equivalently, the Gauss-Newton system of equations associated with Figure~\ref{fig:factor} can be written in the following form:
\begin{equation}
	\underbrace{\left(\mathbf{A}^{-T} \mathbf{Q}^{-1} \mathbf{A} + \mathbf{C}^T \mathbf{R}^{-1} \mathbf{C} \right)}_{\mathbf{L}} \hat{\mathbf{x}} = \underbrace{\mathbf{A}^{-T} \mathbf{Q}^{-1} \check{\mathbf{x}} + \mathbf{C}^T \mathbf{R}^{-1} \mathbf{y}}_{\mathbf{r}},
\end{equation} where $\mathbf{L}$ is block-tridiagonal,
\begin{align}
	\mathbf{L} &= \begin{bmatrix} \mathbf{L}_{0,0} & \mathbf{L}_{0,1:3} \\ \mathbf{L}_{0,1:3}^T & \mathbf{L}_{1:3,1:3} & \mathbf{L}_{1:3,4} \\ & \mathbf{L}_{1:3,4}^T & \mathbf{L}_{4,4} & \mathbf{L}_{4,5:7} \\ & & \mathbf{L}_{4,5:7}^T & \mathbf{L}_{5:7,5:7} & \mathbf{L}_{5:7,8} \\ & & & \mathbf{L}_{5:7,8}^T & \mathbf{L}_{8,8}
	\end{bmatrix} = \mleft[ \begin{array}{c|ccc|c|ccc|c}
		* & * & & & & & & &  \\ \hline
		* & * & * & & & & & &\\
		& * & * & * & & & & &\\
		& &  * & * & * & & &\\ \hline
		& & & * & * & * & &\\ \hline
		& & & & * & * & * & \\
		& & & & & * & * & * \\
		& & & & & & * & * & * \\ \hline
		& & & & & & & * & * \\
	\end{array} \mright].
\end{align} Here, we consider the case where we would like to marginalize the full posterior such that we only retain states $\{\mathbf{x}_0, \mathbf{x}_4, \mathbf{x}_8\}$. After marginalizing out the unwanted states, our system becomes
\begin{equation}
	\mathbf{L}_{\text{small}} \hat{\mathbf{x}}_{\text{small}} = \mathbf{r}_{\text{small}},
\end{equation} where by the Schur complement,
\begin{align}
	\mathbf{L}_{\text{small}} &= \begin{bmatrix}
		\mathbf{L}_{0,0} & & \\ & \mathbf{L}_{4,4} & \\ & & \mathbf{L}_{8,8} \end{bmatrix} - \begin{bmatrix}
		\mathbf{L}_{0,1:3} & \\ \mathbf{L}_{1:3, 4}^T & \mathbf{L}_{4,5:7} \\ & \mathbf{L}_{5:7, 8}^T \end{bmatrix} \begin{bmatrix}
		\mathbf{L}_{1:3,1:3} & \\ & \mathbf{L}_{5:7,5:7} \end{bmatrix} ^{-1} \begin{bmatrix}
		\mathbf{L}_{0,1:3}^T & \mathbf{L}_{1:3,4} & \\ & \mathbf{L}_{4,5:7}^T & \mathbf{L}_{5:7,8} \end{bmatrix} \nonumber \\
	&= \mleft[ \begin{array}{c|c|c}
		* & & \\ \hline & * & \\ \hline & & * \end{array} \mright] - \mleft[ \begin{array}{ccc|ccc}
		* & &  & \\ \hline & & * & * & &  \\ \hline & & & & & * \end{array} \mright] \times \mleft[ \begin{array}{ccc|ccc}
		* & * & \\ * & * & * \\ & * & * \\ \hline & & & * & * \\ & & & * & * & * \\ & & & & * & *  \end{array} \mright]^{-1} \mleft[ \begin{array}{c|c|c} * & & \\ & & \\ & * & \\ \hline & * & \\ & & \\ & & * \end{array} \mright]. \nonumber
\end{align} $\mathbf{L}_{\text{small}}$ can be computed efficiently by exploiting the primary and secondary sparsity. Note that \begin{align}
	&\begin{bmatrix}
		\mathbf{L}_{1:3,1:3} & \\ & \mathbf{L}_{5:7,5:7} \end{bmatrix} ^{-1} \begin{bmatrix}
		\mathbf{L}_{0,1:3}^T & \mathbf{L}_{1:3,4} & \\ & \mathbf{L}_{4,5:7}^T & \mathbf{L}_{5:7,8} \end{bmatrix} = \begin{bmatrix}
		\mathbf{L}_{1:3,1:3} \setminus \mathbf{L}_{0,1:3}^T & \mathbf{L}_{1:3,1:3} \setminus \mathbf{L}_{1:3,4} & \\ & \mathbf{L}_{5:7,5:7} \setminus \mathbf{L}_{4,5:7}^T & \mathbf{L}_{5:7,5:7} \setminus \mathbf{L}_{5:7,8} \end{bmatrix},
\end{align} where each entry similar to $\mathbf{L}_{1:3,1:3} \setminus \mathbf{L}_{1:3,4}$ is shorthand for solving $\mathbf{L}_{1:3,1:3} \boldsymbol{\ell} = \mathbf{L}_{1:3,4}$ for $\boldsymbol{\ell}$. Each of these terms can be solved in linear time thanks to $\mathbf{L}_{1:3,1:3}$ and $\mathbf{L}_{5:7,5:7}$ being block-tridiagonal. Thus, $\mathbf{L}_{\text{small}}$ can be constructed in linear time, the same time complexity as the classic approach. Furthermore, it can be shown that the resulting matrix $\mathbf{L}_{\text{small}}$ is block-tridiagonal. In summary, Schur complement preintegration is a generalization of classic preintegration that can handle both binary motion prior factors and unary measurement factors while retaining the same linear time complexity as classic preintegration.

\section{Analytical Gradients for Training the Singer Prior} \label{app:singer_grad}

Following the method presented by Wong et al. \cite{wong_ral20} for training the parameters of the Singer prior, we found it necessary to derive the analytical gradients of our objective with respect to the desired parameters. Since these gradients were not provided in \cite{wong_ral20}, we provide them here instead for the convenience of the reader. Starting with the objective from \eqref{eq:singer_obj}, the discrete-time covariance $\mathbf{Q}_k$ of the Singer prior can be written as the product of two factors where
\begin{equation}
	\mathbf{Q}_k = \underbrace{\begin{bmatrix}\boldsymbol{\sigma}^2 & & \\ & \boldsymbol{\sigma}^2 & \\ & & \boldsymbol{\sigma}^2 \end{bmatrix}}_{\mathbf{Q}_{\boldsymbol{\sigma}^2}} \mathbf{Q}(\Delta t_k, \boldsymbol{\alpha}).
\end{equation} The components of $\mathbf{Q}(\Delta t_k, \boldsymbol{\alpha})$ are provided by Wong et al. \cite{wong_ral20} and are repeated here,
\begin{equation}
	\mathbf{Q}(\Delta t_k, \boldsymbol{\alpha}) = \begin{bmatrix} \mathbf{Q}_{11} & \mathbf{Q}_{12} & \mathbf{Q}_{13} \\
		\mathbf{Q}_{12}^T & \mathbf{Q}_{22} & \mathbf{Q}_{23} \\ \mathbf{Q}_{13}^T & \mathbf{Q}_{23}^T & \mathbf{Q}_{33} \end{bmatrix},
\end{equation}
where
\begin{subequations}
	\begin{align}
		\mathbf{Q}_{11} &= \frac{1}{2} \a^{-5} \Big( \mathbf{1} - e^{-2\a \dt_k} + 2 \a \dt_k + \frac{2}{3} \a^3\dt^3_k - 2 \a^2 \dt^2_k - 4\a\dt_k e^{-\a \dt_k} \Big),\\
		\mathbf{Q}_{12} &=\frac{1}{2} \a^{-4} \Big( e^{-2\a \dt_k} + \mathbf{1} - 2e^{-\a \dt_k} + 2 \a \dt_k e^{-\a \dt_k} - 2\a\dt_k + \a^2\dt_k^2 \Big), \\
		\mathbf{Q}_{13} &=\frac{1}{2} \a^{-3} \left(\mathbf{1} - e^{-2\a \dt_k} -2\a\dt_ke^{-\a \dt_k} \right), \\
		\mathbf{Q}_{22} &=\frac{1}{2} \a^{-3} \left(4 e^{-\a \dt_k} - 3\cdot\mathbf{1} - e^{-2\a \dt_k} + 2\a\dt_k \right), \\
		\mathbf{Q}_{23} &=\frac{1}{2} \a^{-2} \left( e^{-2\a \dt_k} + \mathbf{1} - 2 e^{-\a \dt_k}  \right), \\
		\mathbf{Q}_{33} &=\frac{1}{2} \a^{-1} \left( \mathbf{1} -  e^{-2\a \dt_k} \right).
	\end{align}
\end{subequations} The motion error is given by
\begin{equation}
	\mathbf{e}_k = \mathbf{x}_k - \p(t_k, t_{k-1}) \mathbf{x}_{k-1},
\end{equation} where
\begin{align}
	&\p(t_k, t_{k-1}) = \begin{bmatrix} \mathbf{1} & \Delta t_k \mathbf{1} & (\boldsymbol{\alpha} \Delta t_k - \mathbf{1} + \exp(-\boldsymbol{\alpha} \Delta t_k))\boldsymbol{\alpha}^{-2} \\
		\mathbf{0} & \mathbf{1} & (\mathbf{1} -\exp(-\boldsymbol{\alpha} \Delta t_k )  ) \boldsymbol{\alpha}^{-1} \\ \mathbf{0} & \mathbf{0} & \exp(-\boldsymbol{\alpha} \Delta t_k ) \end{bmatrix}
\end{align} is the state transition function. $\boldsymbol{\sigma}^2$ and $\a$ are both diagonal matrices, whose size depends on the dimension of the state. For example, for a 6D state, $\boldsymbol{\sigma}^2 = \text{diag}(\sigma^2_1, \sigma^2_2, \sigma^2_3, \sigma^2_4, \sigma^2_5, \sigma^2_6)$. The gradients of the objective with respect to the components of $\boldsymbol{\sigma}^2$ and $\a$ are then
\begin{subequations}
	\begin{align}
		\frac{\partial J_t}{\partial \alpha_i} &= \frac{1}{2} \sum_k \left\{ 2 \mathbf{e}_k^T \mathbf{Q}_k^{-1} \frac{\partial \mathbf{e}_k}{\partial \alpha_i} - \mathbf{e}_k^T \mathbf{Q}_k^{-1} \frac{\partial \mathbf{Q}_k}{\partial \alpha_i} \mathbf{Q}_k^{-1}  \mathbf{e}_k +~\text{tr} \left(\mathbf{Q}_k^{-1}  \frac{\partial \mathbf{Q}_k}{\partial \alpha_i} \right) \right\}, \\
		\frac{\partial J_t}{\partial \sigma^2_i} &=   \frac{3K}{2 \sigma^2_i} - \frac{1}{2} \sum_k \frac{1}{\sigma^4_i} \mathbf{e}_k^T \mathbf{Q}(\Delta t_k, \boldsymbol{\alpha})^{-1} \frac{\partial \mathbf{Q}_{\boldsymbol{\sigma}^2}}{\partial \sigma^2_i} \mathbf{e}_k,
	\end{align}
\end{subequations} for each $\alpha_i$ and $\sigma^2_i$, respectively, where $J = \sum_{t=1}^T J_t$. The partial derivatives of $\mathbf{Q}(\Delta t_k, \boldsymbol{\alpha})$ and $\mathbf{e}_k$ with respect to $\alpha_i$ are then given by
\begin{subequations}
\begin{align}
	\frac{\partial \q_{11}}{\partial \alpha_i} &= \left[-\frac{2 \dt_k^3}{3 \alpha_i^3} + \frac{\dt_k^2 (2 \expon + 3)}{\alpha_i^4} + \frac{5 (\expontwo - 1)}{2 \alpha_i^6} + \frac{\dt_k(\expontwo + 8 \expon - 4)}{\alpha_i^5} \right] \boldsymbol{\delta}_{ii}, \\
	\frac{\partial \q_{12}}{\partial \alpha_i} &= \left[-\frac{\dt_k^2 (\expon + 1)}{\alpha_i^3} +  \frac{\dt_k (3 -  \expontwo - 2 \expon)}{\alpha_i^4} + \frac{4 \expon - 2 \expontwo - 2}{\alpha_i^5} \right] \boldsymbol{\delta}_{ii},\\
	\frac{\partial \q_{13}}{\partial \alpha_i} &= \left[\frac{\dt_k^2 \expon}{\alpha_i^2} + \frac{3 (\expontwo - 1)}{2 \alpha_i^4} + \frac{\dt_k( \expontwo + 2 \expon)}{\alpha_i^3} \right] \boldsymbol{\delta}_{ii}, \nonumber\\
	\frac{\partial \q_{22}}{\partial \alpha_i} &= \left[\frac{3 \expontwo - 12 \expon + 9}{2 \alpha_i^4} + \frac{\dt_k( \expontwo - 2 \expon - 2)}{\alpha_i^3} \right] \boldsymbol{\delta}_{ii}, \\
	\frac{\partial \q_{23}}{\partial \alpha_i} &=  \left[\frac{2 \expon - \expontwo - 1}{\alpha_i^3} + \frac{\dt_k(  \expon - \expontwo)}{\alpha_i^2} \right] \boldsymbol{\delta}_{ii}, \\
	\frac{\partial \q_{33}}{\partial \alpha_i} &= \left[\frac{\expontwo - 1}{2 \alpha_i^2} + \frac{\dt_k \expontwo}{\alpha_i}\right] \boldsymbol{\delta}_{ii}, \\
	\frac{\partial \mathbf{e}_k}{\partial \alpha_i} = &- \begin{bmatrix} \left(\frac{2(1 -  \expon)}{\alpha_i^3} - \frac{\dt_k (\expon + 1)}{\alpha_i^2}\right) \boldsymbol{\delta}_{ii} \\ \left(\frac{\expon - 1}{\alpha_i^2} + \frac{\dt_k \expon}{\alpha_i} \right) \boldsymbol{\delta}_{ii} \\ \left(-\dt_k \expon \right) \boldsymbol{\delta}_{ii}\end{bmatrix} \begin{bmatrix} \mathbf{0} & \mathbf{0} & \mathbf{1} \end{bmatrix} \mathbf{x}_k,
\end{align}
\end{subequations} where $\boldsymbol{\delta}_{ii}$ is the Kronecker delta. We can use these gradients to learn the parameters of the Singer prior using gradient descent. In order to speed up training, we can solve for the optimal value of $\sigma_i^2$ at each iteration of gradient descent:
\begin{equation}
	\sigma_i^{2^\star} = \frac{1}{3KT} \sum_t\sum_k \mathbf{e}_{k,t}^T \mathbf{Q}(\Delta t_{k,t}, \boldsymbol{\alpha})^{-1} \frac{\partial \mathbf{Q}_{\boldsymbol{\sigma}^2}}{\partial \sigma^2_i} \mathbf{e}_{k,t}.
\end{equation} Note that $\mathbf{Q}_k$ is numerically unstable for $\alpha < 1.0$. In this case, we use a Taylor series expansion about $\alpha = 0$ as an approximation. The Jacobian $\frac{\partial \mathbf{Q}_k}{\partial \alpha_i}$ is also numerically unstable for $\alpha < 4.0$. In this case, we can approximate the components of this matrix with either a Laurent series or Taylor series as $\alpha \rightarrow 0$.

The previous gradients work well for learning the parameters of a Gaussian process in simulation where the ground truth measurements of the state are noiseless. In reality, our source of ground truth will have some measurement covariance that may be estimated or taken from the datasheet of the sensor being used. In this case, computing the gradients of the objective with respect to the components of $\boldsymbol{\sigma}^2$ and $\a$ is slightly more involved. We follow the approach presented by Wong et al. \cite{wong_ral20}. Now, our objective function looks at the entire trajectory at once,
\begin{equation}
	J = -\ln p(\mathbf{y} | \boldsymbol{\sigma}, \boldsymbol{\alpha}) = \frac{1}{2} \mathbf{e}^T \mathbf{Q}^{-1} \mathbf{e} + \frac{1}{2} \ln | \mathbf{Q} | + \frac{n}{2} \ln 2 \pi,
\end{equation} where $\mathbf{e}$ is a stacked version of all the individual error terms from each timestep $\mathbf{e}_k$, and
\begin{equation}
	\mathbf{Q} = \begin{bmatrix} \boldsymbol{\Sigma}_{0,0} & \boldsymbol{\Sigma}_{0,1} \\
		\boldsymbol{\Sigma}_{1,0}^T & \boldsymbol{\Sigma}_{1,1} & \boldsymbol{\Sigma}_{1,2} \\
		& \boldsymbol{\Sigma}_{1,2}^T & \ddots & \ddots \\
		& & \ddots & \boldsymbol{\Sigma}_{K, K}
	\end{bmatrix},
\end{equation} where
\begin{subequations}
\begin{align}
	&\boldsymbol{\Sigma}_{k,k} \approx \mathbf{R}_{k} + \boldsymbol{\Phi}(t_{k}, t_{k-1}) \mathbf{R}_{k-1} \boldsymbol{\Phi}(t_{k}, t_{k-1})^T + \mathbf{Q}_k, \\
	&\boldsymbol{\Sigma}_{k, k+1} \approx - \mathbf{R}_{k} \boldsymbol{\Phi}(t_{k+1}, t_{k})^T,
\end{align}
\end{subequations} and $\mathbf{R}_k$ is the measurement covariance associated with local variable $\mathbf{x}_k$. The gradient of the objective function $J$ with respect to GP parameter $\theta$ is
\begin{equation} \label{eq:gp_gradient}
	\frac{\partial J}{\partial \theta} = -\frac{1}{2} \mathbf{e}^T \mathbf{Q}^{-1} \frac{\partial \mathbf{Q}}{\partial \theta} \mathbf{Q}^{-1} \mathbf{e} + \mathbf{e}^T \mathbf{Q}^{-1} \frac{\partial \mathbf{e}}{\partial \theta} + \frac{1}{2} \text{tr} \left(\mathbf{Q}^{-1} \frac{\partial \mathbf{Q}}{\partial \theta} \right)
\end{equation} where each of the Jacobians is evaluated using the current value of $\theta$. We can compute the trace in $O(K)$ time by first computing only the block-tridiagonal components of $\mathbf{Q}^{-1}$. Since $\mathbf{Q}$ is itself block-tridiagonal, we can compute the blocks of the inverse that we need in $O(K)$ time \cite{barfoot_ser17}. Then, we can compute the trace of the matrix product in $O(K)$ time by only computing the elements along the diagonal of the matrix product. $\mathbf{Q}^{-1} \mathbf{e}$ can also be evaluated in $O(K)$ time by solving $\mathbf{Q} \mathbf{x} = \mathbf{e}$ for $\mathbf{x}$ using a sparse Cholesky solver. Using the gradient in \eqref{eq:gp_gradient} for each parameter, we can learn the parameters of the Gaussian process by minimizing the negative log likelihood using gradient descent.

\change{
\section{IMU-as-Input Lidar-Inertial Baseline Jacobians} \label{app:baseline_jacobians}

Perturbations to the state variables are defined as $\mathbf{C}_{iv} = \overline{\mathbf{C}}_{iv}\exp(\delta \phi^\wedge)$, $\mathbf{r}_i^{vi} = \bar{\mathbf{r}}_i^{vi} + \overline{\mathbf{C}}_{iv} \delta \mathbf{r}$, $\mathbf{v}_i^{vi} = \bar{\mathbf{v}}_i^{vi} + \overline{\mathbf{C}}_{iv} \delta \mathbf{v}$, $\mathbf{b} = \bar{\mathbf{b}} + \delta \mathbf{b}$. The Jacobians of the point-to-plane error function \eqref{eq:baseline_error} with respect to perturbations to the state variables are provided here,

\begin{equation}
	\frac{\partial \mathbf{e}_j}{\partial \delta \mathbf{x}} =  \begin{bmatrix} \frac{\partial \mathbf{e}_j}{\partial  \mathbf{r}^{vi}_i(\tau_j)} & \frac{\partial \mathbf{e}_j}{\partial \delta \mathbf{C}_{iv}(\tau_j)} \end{bmatrix} \times \begin{bmatrix}  \frac{\partial \mathbf{r}_i^{vi}(\tau_j)}{\partial \mathbf{r}_\ell} \frac{\partial \mathbf{r}_{\ell}}{\partial \delta \mathbf{x}} + \frac{\partial \mathbf{r}_i^{vi}(\tau_j)}{\partial \mathbf{r}_{\ell+1}} \frac{\partial \mathbf{r}_{\ell + 1}}{\partial \delta \mathbf{x}} \\ \frac{\partial \delta \mathbf{C}_{iv}(\tau_j)}{\partial \delta \mathbf{C}_\ell} \frac{\partial \delta \mathbf{C}_{\ell}}{\partial \delta \mathbf{x}} + \frac{\partial \delta \mathbf{C}_{iv}(\tau_j)}{\partial \delta \mathbf{C}_{\ell+1}} \frac{\partial \delta \mathbf{C}_{\ell+1} }{\partial \delta \mathbf{x}} \end{bmatrix},
\end{equation}
where
\begin{subequations}
\begin{align}
\frac{\partial \mathbf{e}_j}{\partial  \mathbf{r}^{vi}_i(\tau_j)} &= -\mathbf{n}_j^T, \quad \frac{\partial \mathbf{e}_j}{\partial \delta \mathbf{C}_{iv}(\tau_j)} = \mathbf{n}_j^T \left( \overline{\mathbf{C}}_{iv}(\tau_j)(\mathbf{C}_{vs}\mathbf{q}_j +\mathbf{r}_v^{sv})^\wedge \right), \\
 \frac{\partial \mathbf{r}_i^{vi}(\tau_j)}{\partial \mathbf{r}_\ell} &= (1 - \alpha) \mathbf{1}, \quad   \frac{\partial \mathbf{r}_i^{vi}(\tau_j)}{\partial \mathbf{r}_{\ell+1}} = \alpha \mathbf{1}, \\
  \frac{\partial \delta \mathbf{C}_{iv}(\tau_j)}{\partial \delta \mathbf{C}_\ell} &= \mathbf{1} - \mathbf{A}(\alpha, \boldsymbol{\phi}),   \quad  \frac{\partial \delta \mathbf{C}_{iv}(\tau_j)}{\partial \delta \mathbf{C}_{\ell+1}} =  \mathbf{A}(\alpha, \boldsymbol{\phi}),
\end{align}
\end{subequations} where $\mathbf{A}(\alpha, \boldsymbol{\phi}) = \alpha \mathbf{J}_r(\alpha \boldsymbol{\phi}) \mathbf{J}_r(\boldsymbol{\phi})^{-1}$, $\boldsymbol{\phi} =  \ln(\mathbf{C}_\ell^T \mathbf{C}_{\ell+1})^\vee$, and

\begin{subequations}
	\begin{align}
\frac{\partial \delta \mathbf{C}_\ell}{\partial \delta \mathbf{C}_i} &= \Delta \overline{\mathbf{C}}_{i\ell}^T,~\text{where}\quad  \Delta \mathbf{C}_{i\ell} = \prod_{k=i}^{\ell-1} \exp\left(\Delta t_k  (\tilde{\boldsymbol{\omega}}_k - \mathbf{b}_\omega(t_k))^\wedge \right ),\\
\frac{\partial \delta \mathbf{C}_\ell}{\partial \delta \mathbf{b}_\omega(t_i)} &= -\sum_{k=i}^{\ell-1} \Delta \overline{\mathbf{C}}_{k+1,\ell}^T \mathbf{J}_r(\boldsymbol{\phi}_k) \Delta t_k, ~\text{where}\quad \boldsymbol{\phi}_k = \Delta t_k  (\tilde{\boldsymbol{\omega}}_k - \mathbf{b}_\omega(t_k)),\\
\frac{\partial\mathbf{v}_\ell}{\partial \delta \mathbf{v}_i} &= \mathbf{C}_i, \\
\frac{\partial \mathbf{v}_\ell}{\partial \delta \mathbf{C}_i} &= -\sum_{k=i}^{\ell-1}  \mathbf{C}_k (\tilde{\mathbf{a}}_k - \mathbf{b}_a(t_k))^\wedge \Delta \overline{\mathbf{C}}_{ik}^T \Delta t_k, \\
\frac{\partial \mathbf{v}_\ell}{\partial \delta \mathbf{b}_\omega(t_i)} &= -\sum_{k=i}^{\ell-1}  \mathbf{C}_k (\tilde{\mathbf{a}}_k - \mathbf{b}_a(t_k))^\wedge \frac{\partial \delta \mathbf{C}_k}{\partial \delta \mathbf{b}_\omega(t_i)} \Delta t_k, \\
\frac{\partial \mathbf{v}_\ell}{\partial \delta \mathbf{b}_a(t_i)} &= -\sum_{k=i}^{\ell-1} \mathbf{C}_k \Delta t_k, \\
\frac{\partial \mathbf{r}_\ell}{\partial \delta \mathbf{v}_i} &= \mathbf{C}_i \Delta t_{ij}, \\
\frac{\partial \mathbf{r}_\ell}{\partial \delta \mathbf{C}_i} &= \sum_{k=i}^{\ell-1} \left[ \frac{\partial  \mathbf{v}_k}{\partial \delta \mathbf{C}_i} \Delta t_k - \frac{1}{2}  \mathbf{C}_k (\tilde{\mathbf{a}}_k - \mathbf{b}_a(t_k))^\wedge \Delta \overline{\mathbf{C}}_{ik}^T \Delta t_k^2 \right], \\
\frac{\partial \mathbf{r}_\ell}{\partial \delta \mathbf{b}_\omega(t_i)} &= \sum_{k=i}^{\ell-1} \left[ \frac{\partial \mathbf{v}_k}{\partial \delta \mathbf{b}_\omega (t_i) } \Delta t_k - \frac{1}{2}  \mathbf{C}_k (\tilde{\mathbf{a}}_k - \mathbf{b}_a(t_k))^\wedge \frac{\partial \delta \mathbf{C}_k}{\partial \delta \mathbf{b}_\omega(t_i)} \Delta t_k^2 \right], \\
\frac{\partial \mathbf{r}_\ell}{\partial \delta \mathbf{b}_a (t_i)} &= \sum_{k=i}^{\ell-1} \left[ \frac{\partial \mathbf{v}_k}{\partial \delta \mathbf{b}_a (t_i) } \Delta t_k - \frac{1}{2} \mathbf{C}_k \Delta t_k^2 \right]. \\
	\end{align}
\end{subequations}

}

%\textbf{**Note:} This method of looking at the entire trajectory at once only works for linear (vector space) state spaces, for Lie groups we have a sequence of local GPs, so we should be summing over terms and not include the cross-covariance between ``knots".
%
%\textbf{**Note 2:} Are we really learning the prior that best matches the underlying process noise or are we learning the prior that was used by the source of ground truth? For example, the SBET might simply be using a constant velocity motion prior. It might be best to simply use ESGVI for this problem.
%\newpage

\else

\ack{Author Contributions}{\fontsize{8}{10}\selectfont Keenan Burnett carried out the research and wrote the article, Angela Schoellig provided feedback on research, Tim Barfoot conceived of the study and provided feedback on research.}

\ack{Financial Support} {\fontsize{8}{10}\selectfont This work was supported in part by the National Sciences and Engineering Research Council of Canada (NSERC) and by an Ontario Research Fund: Research Excellence (ORF-RE) grant.}

\ack{Competing Interests} {\fontsize{8}{10}\selectfont The authors declare none.}

\ack{Data Availability} {\fontsize{8}{10}\selectfont Data sharing is not applicable to this article as no new data were created or analyzed in this study.}

\fi

\bibliography{bib/references}
%\balance

\end{document}